\documentclass[5p,times,twocolumn]{elsarticle}
\usepackage{amsmath,amsfonts,amssymb}
\usepackage{graphicx}
\usepackage{subcaption}
\usepackage{epstopdf}
\usepackage{algorithm}
\usepackage{algorithmic}
\usepackage{array}
\usepackage{multirow}
\usepackage{tabularx}
\usepackage{xcolor}
\usepackage{url}

\newcommand{\figref}[1]{Fig.~\ref{#1}}
\newcommand{\R}{\mathbb{R}}

\journal{Robotics and Autonomous Systems}
\begin{document}
\begin{frontmatter}
\title{Dual-Arm Box Manipulation via Online Inertial Estimation and Convex Wrench Optimization}
\author[ntu]{Kenzhi Iskandar Wong\fnref{equal}}
\author[ntu]{Lin Yang\fnref{equal}}
\author[ntu]{Qian Ying Lee}
\author[ntu]{Domenico Campolo\corref{cor1}}
\ead{d.campolo@ntu.edu.sg}
\affiliation[ntu]{
organization={School of Mechanical and Aerospace Engineering, Nanyang Technological University}, country={Singapore} }
\fntext[equal]{These authors contributed equally to this work.}
\cortext[cor1]{Corresponding author.}
\begin{abstract}
Dual-arm robotic manipulation often requires lifting and transporting objects whose mass and center of mass are not known in advance. These uncertainties affect the force-moment balance required for stable lifting, and improper regulation of contact wrenches can lead to slip, object drop, orientation deviation, or excessive squeezing. This paper presents a friction-aware dual-arm box-manipulation framework for objects with unknown inertial properties. The framework combines online inertial estimation with convex contact-wrench optimization. The inertial estimation module estimates the object mass and center of mass from measured contact wrenches, while the optimization module computes friction-feasible contact forces and torsional moments through a second-order cone program (SOCP) under ellipsoidal friction-limit-surface constraints. An offline trajectory refinement stage is also included to reduce undesired object-environment contact when geometric constraints are present. By enforcing friction feasibility as a hard constraint and minimizing contact effort within the feasible region, the framework achieves stable lifting without treating slip avoidance and excessive squeezing as separately tuned objectives. Experiments on a real dual-arm robotic system under different center-of-mass configurations demonstrate that the method lifts objects with unknown inertial properties while maintaining stable frictional contact.
\end{abstract}


\end{frontmatter}

\section{Introduction}\label{sec_introduction}

\begin{figure}
    \centering
    \includegraphics[width=\columnwidth]{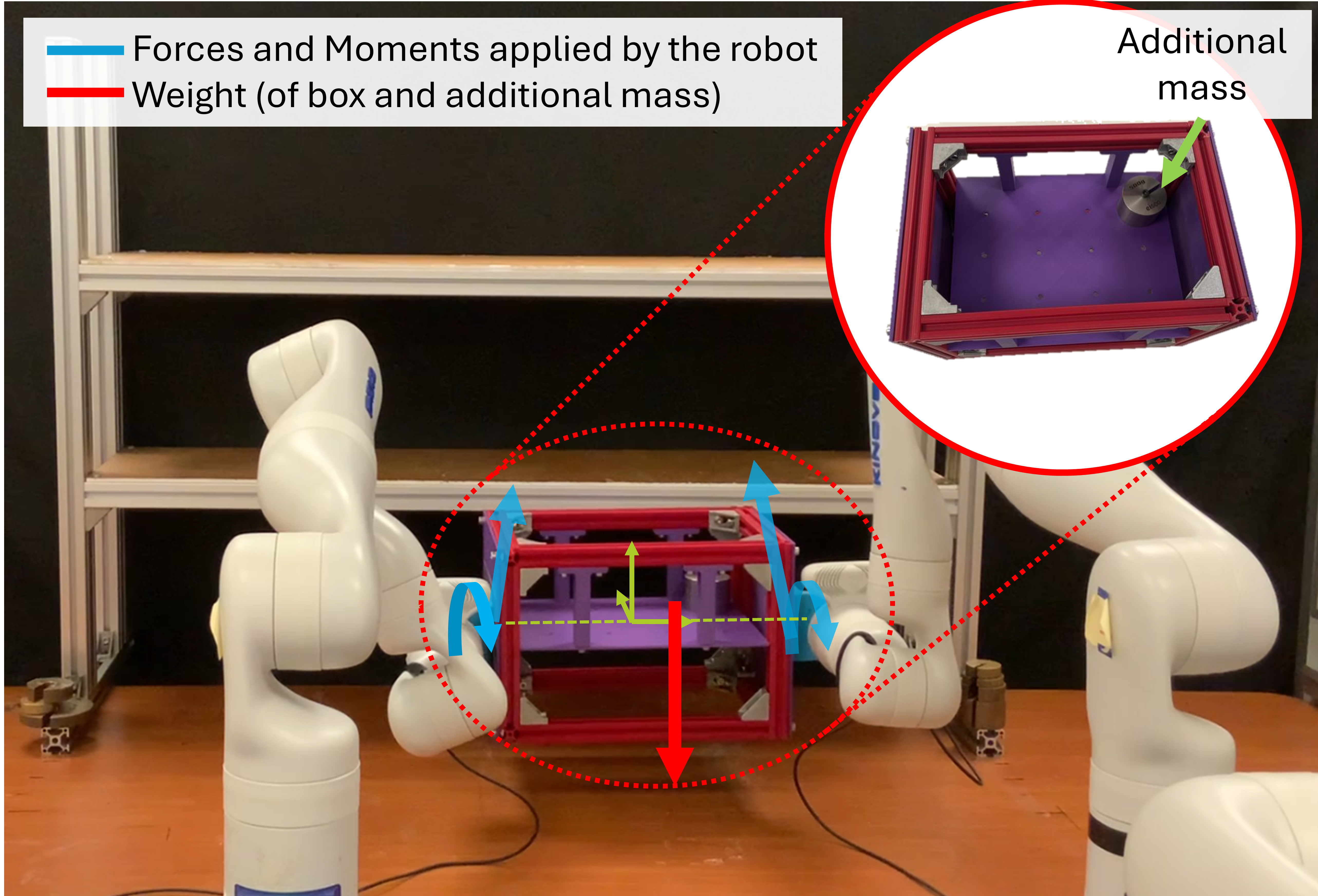}
    \caption{Dual-arm box manipulation with initially unknown mass and center of mass. Contact measurements are used to estimate these properties, and friction-constrained wrench optimization determines the contact forces and moments for lifting and transport}.
    \label{fig:overview}
\end{figure}

Boxes with similar external geometry can have different masses and centers of mass (CoM) because their contents are distributed differently. These uncertainties complicate robotic handling in logistics and warehouse environments, where coordinated dual-arm manipulation is useful for objects that cannot be handled conveniently by a single gripper \cite{smith2012dual,correll2016analysis}. In the setting considered here, a box is supported through frictional contacts on two opposite sides, as illustrated in \figref{fig:overview}. The contact wrenches must support the object's weight and balance the moments caused by its mass distribution. Improper regulation of these wrenches can lead to slip, object drop, orientation deviation, or excessive squeezing. Stable manipulation therefore requires sufficient contact forces and moments to maintain the grasp without applying unnecessarily large squeezing forces.

\begin{figure}
    \centering
    \includegraphics[width=\columnwidth]{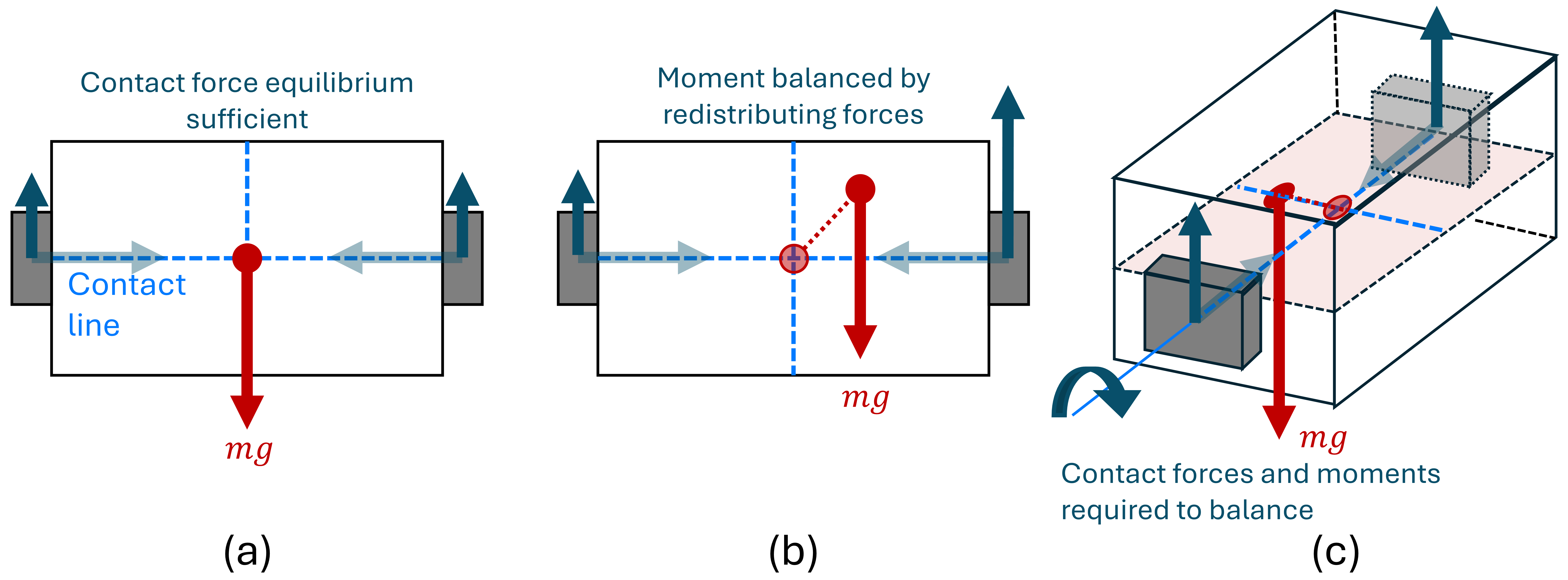}
    \caption{Dual-arm lifting with known CoM locations. (a) A centered CoM permits symmetric load sharing. (b) An in-plane offset can be compensated by redistributing the contact forces. (c) Contact moments are needed when the gravitational moment has a component along the line joining the contacts.}
    \label{fig:box_three_cases}
\end{figure}

The role of contact moments is illustrated in \figref{fig:box_three_cases}. For a centered CoM, the load can be shared symmetrically between the contacts, as shown in \figref{fig:box_three_cases}(a). An offset within the lifting plane can be accommodated by redistributing the supporting forces, as shown in \figref{fig:box_three_cases}(b). When the CoM offset produces a gravitational moment along the line joining the contacts, supporting forces alone cannot balance this moment, and contact moments are required, as shown in \figref{fig:box_three_cases}(c). Stable lifting therefore depends on both the distribution of contact wrenches and their frictional feasibility \cite{kerr1986analysis,bicchi2000robotic}.

The need to coordinate object motion and contact forces has motivated a range of model-based approaches to cooperative manipulation. In these approaches, the relationship between robot motion, object motion, and contact forces is used to guide the actions of the manipulators. Caccavale et al. developed a dual-arm impedance controller that combines compliant behavior at the object level with the regulation of internal loading at the end effectors \cite{caccavale2008six}. This allows the manipulators to respond to external interaction while avoiding unnecessarily large forces on the grasped object. Erhart et al. investigated related challenges in cooperative mobile manipulation, where uncertainties in robot and object geometry can cause inconsistencies between the commanded motions of the manipulators \cite{erhart2013impedance}. Their impedance-based coordination scheme limits the undesired internal forces arising from these inconsistencies. For large objects supported on a planar surface, Koutras et al. proposed a passivity-based bimanual controller that moves and rotates the object without requiring knowledge of its dynamics or the friction dynamics at the supporting surface \cite{koutras2023towards}. This extends coordinated manipulation to nonprehensile tasks in which support from the environment contributes to object handling. Optimization-based force allocation has also been studied: Boyd and Wegbreit formulated the computation of contact forces that balance a known external wrench under friction constraints as a convex optimization problem \cite{boyd2007fast}. Beyond transport, Bombile and Billard combined coordinated motion with contact-wrench optimization for fast grabbing and tossing \cite{bombile2022dual}. Their approach addresses the changing interaction conditions as the arms establish contact, move the object, and release it. Together, these studies illustrate how coordinated motion and force regulation support cooperative manipulation, from compliant transport to dynamic object handling. However, when the box mass and CoM are initially unknown, the load that the two arms need to support must first be estimated. The challenge is to use these estimates to determine contact forces and moments that balance the box while remaining within the friction limits.

Alongside these model-based methods, learning-based approaches have been developed to guide manipulation from visual and tactile observations. Rather than specifying every relationship between sensory feedback and robot action analytically, these approaches learn useful behavior from demonstrations or repeated interaction. Levine et al. learned visual hand-eye coordination from large-scale grasping data \cite{levine2018learning}. Their method predicts grasp success from camera images and candidate gripper motions, allowing the robot to adjust its movement continuously as the scene changes. The resulting feedback behavior enables the robot to correct unsuccessful approaches and grasp previously unseen objects. For bimanual interaction, Lin et al. introduced Bi-Touch, which uses deep reinforcement learning to acquire coordinated manipulation policies from tactile feedback \cite{lin2023bi}. Policies are trained in simulation and transferred to a physical dual-arm system for tasks including pushing, reorienting, and gathering objects. Experiments with unseen objects and external perturbations demonstrate the ability of tactile feedback to support adjustments during contact. Beyond learning manipulation policies, Dutta et al. combined visual and tactile observations with learned interaction models to infer object properties through exploratory pushing and pulling \cite{dutta2025predictive}. Their framework selects informative actions, showing how physical interaction can reveal properties that are not directly available from visual observations alone. These methods show how sensory feedback can guide manipulation and reveal object properties through interaction. However, translating this information into contact forces and moments that explicitly satisfy equilibrium and friction constraints is a further step. For dual-arm lifting, this connection is needed to use the estimated loading to maintain a stable grasp without excessive squeezing.

This work addresses that connection for quasi-static box manipulation through opposing frictional contacts. The proposed framework combines offline trajectory refinement, post-lift estimation of mass and CoM components, and friction-constrained contact-wrench optimization. The reference trajectory is first refined offline when required to avoid surrounding obstacles. After an initial lift, contact measurements are used to estimate the mass and CoM components. These estimates guide the allocation of contact forces and moments, which are then executed through impedance control with force feedback. The contribution lies in integrating these stages to address slip, object drop, and orientation deviation while reducing unnecessary squeezing, and in evaluating the resulting performance through simulation and real-robot experiments. The main contributions are:
\begin{itemize}
    \item An integrated framework combines trajectory refinement, post-lift estimation of mass and CoM, and contact-wrench execution for quasi-static dual-arm manipulation of boxes with unknown internal loading.
    \item An Second-Order Cone Program formulation uses the estimated loading to determine the contact forces and torsional moments that balance the box, minimizing weighted contact effort while satisfying coupled friction constraints.
    \item A safety margin is defined through a Pontryagin difference, contracting the friction-feasible region against bounded wrench perturbations at a fixed normal force.
\end{itemize}

\section{Methods} \label{sec:methods}

\begin{figure*}
    \centering
    \includegraphics[width=\textwidth]{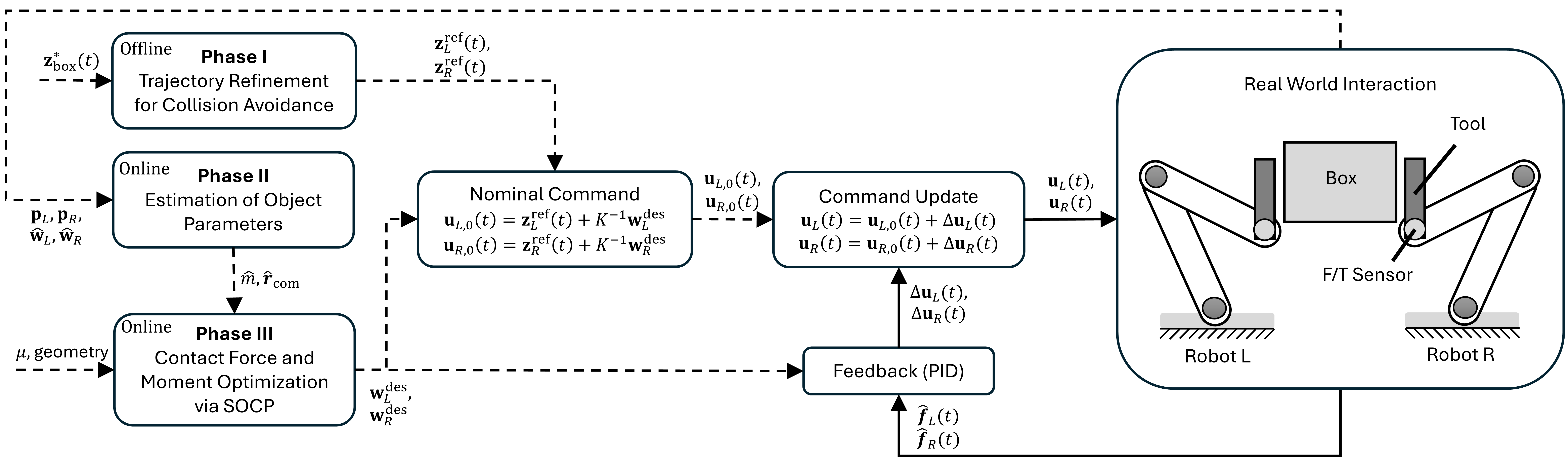}
    \caption{Overview of the proposed framework. An offline trajectory refinement stage is followed by online object parameter estimation and contact wrench optimization. The optimized wrenches are executed through command synthesis and wrench correction in closed-loop interaction with the physical system.}
    \label{fig:framework}
\end{figure*}

Dual-arm manipulation of a rigid object through two frictional contacts is considered. The object is lifted quasi-statically along a Cartesian trajectory obtained from demonstration \cite{lee2025generalizingrobottrajectoriessinglecontext} or prior planning. While the desired object trajectory is known, the object mass and center of mass (CoM) are initially unknown. The objective is to estimate these parameters online and compute friction-feasible contact forces and moments that enable slip-free and drop-free lifting without excessive squeezing.

As summarized in \figref{fig:framework}, the proposed framework consists of three phases and a wrench execution layer:
\begin{itemize}
    \item \textbf{Phase I (offline)} refines the nominal object trajectory offline to avoid undesired object-environment contact while remaining close to the reference motion.
    \item \textbf{Phase II (online)} estimates the object mass and CoM online from measured contact wrenches and robot kinematics.
    \item \textbf{Phase III (online)} computes the contact forces and moments by solving a convex optimization problem subject to force-moment equilibrium and friction constraints.
\end{itemize}
The resulting wrenches are executed through Cartesian impedance control with wrench feedback during lifting.

\subsection{Phase I: Trajectory Refinement for Collision Avoidance} \label{sec:methods_phase1}

\begin{figure}
    \centering
    \includegraphics[width=\columnwidth]{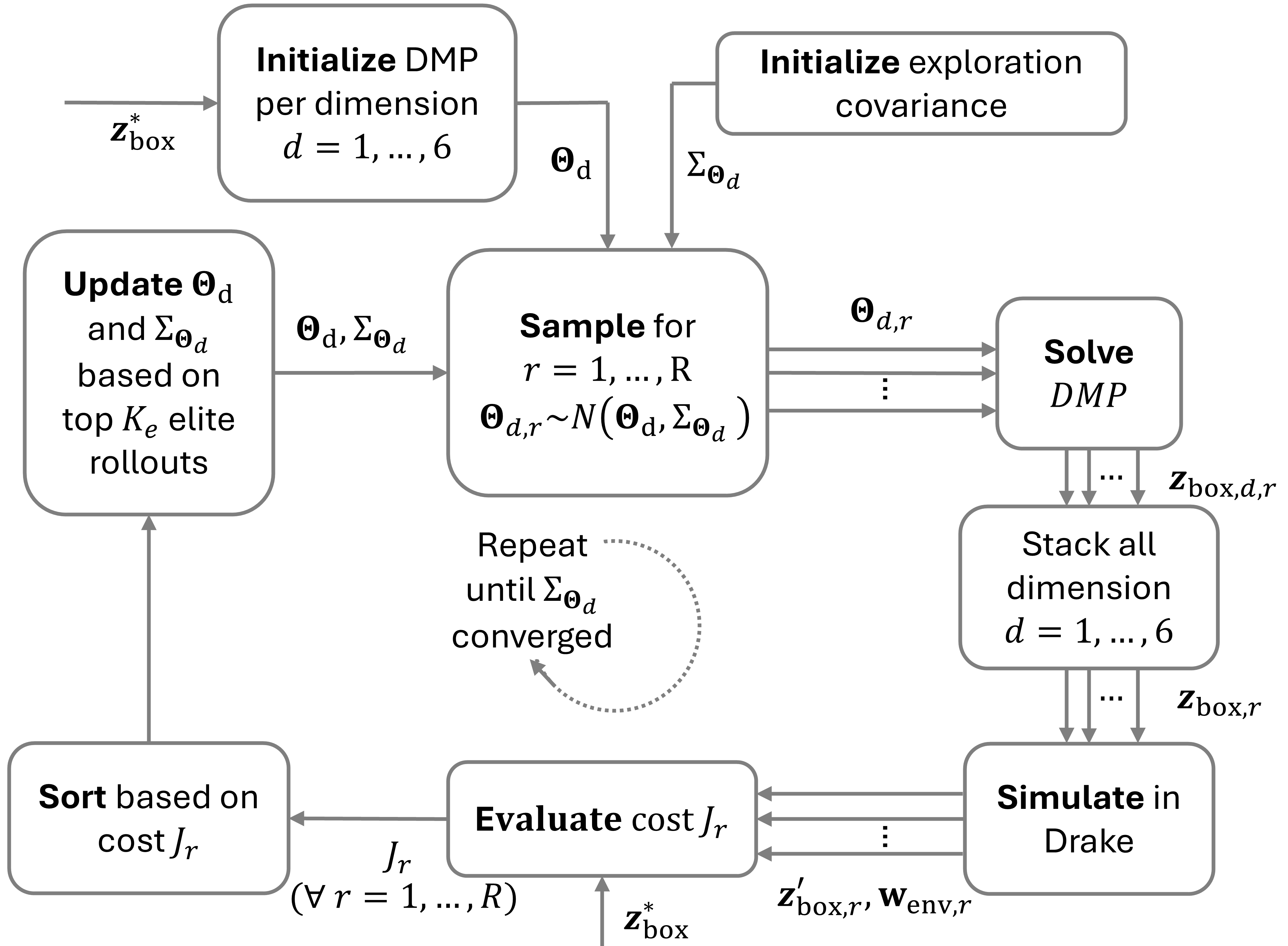}
    \caption{Phase I trajectory refinement via black-box optimization. The objective is to refine a given object trajectory by minimizing object-environment contact while remaining close to the reference trajectory. Optimization is done using DMP-based black-box optimization and physics-based simulation.}
    \label{fig:phase1_optimization}
\end{figure}

Phase I refines the object trajectory offline to reduce undesired object-environment contact while remaining close to the reference motion \cite{yang2025planning}. Let $\mathbf{z}_\text{box}^*(t) \in\mathbb{R}^6$ denote the object pose reference trajectory in local Cartesian pose coordinates. If the trajectory is obtained from demonstration \cite{lee2025generalizingrobottrajectoriessinglecontext}, an associated covariance $\Sigma_{\mathbf{z}_\text{box}^*}(t)\in\mathbb{R}^{6\times 6}$ may also be available. The trajectory is represented using Dynamical Movement Primitives (DMPs) \cite{saveriano2023dynamic}. For each dimension $d=1,\ldots,6$, the DMP is parameterized by weights $\mathbf{\Theta}_d\in\mathbb{R}^{N_{\mathrm{b}}}$, where $N_{\mathrm{b}}$ is the number of basis functions. The weights are initialized by locally weighted regression on the reference trajectory \cite{ijspeert2013dynamical}. The exploration covariance for each trajectory dimension is initialized as
\begin{equation}\label{eq:sigma_init}
    \Sigma_{\mathbf{\Theta}_d}
    = c_d\,\mathbf I_{N_{\mathrm b}},
    \qquad d=1,\ldots,6,
\end{equation}
where $c_d\geq0$ controls the initial exploration magnitude for dimension $d$. Setting $c_d=0$ keeps the corresponding DMP parameters fixed at their reference values. At each iteration, $R$ candidate parameter vectors are sampled as
\begin{equation} \label{eq:sample}
    \mathbf{\Theta}_{d,r} \sim
    \mathcal{N}\!\big(\mathbf{\Theta}_{d},\,\Sigma_{\mathbf{\Theta}_d} \big),
    \quad r=1,\ldots,R .
\end{equation}
Solving the sampled DMPs produces per-dimension candidate trajectories $\mathbf{z}_{\text{box},d,r}(t) \in \R$, which are stacked to form the full candidate object trajectory $\mathbf{z}_{\text{box},r}(t) \in \R^6$. Each candidate is simulated in Drake \cite{drake}, producing the simulated object trajectory $\mathbf{z}_{\text{box},r}'(t) \in \R^6$, and object-environment contact wrenches $\mathbf{w}_{\text{env},r}(t) \in \R^6$:
\begin{equation} \label{eq:drake}
    \big(\mathbf{z}_{\text{box},r}'(t),\, \mathbf{w}_{\text{env},r}(t)\big)
    \gets \textbf{Drake}\big(\mathbf{z}_{\text{box},r}(t)\big).
\end{equation}
Each rollout is evaluated using
\begin{equation} \label{eq:costJ}
    J_r = \alpha J_{1,r} + (1-\alpha) J_{2,r},
\end{equation}
where $\alpha\in(0,1)$ balances tracking of the reference trajectory and avoidance of object-environment contact. The tracking and contact costs are
\begin{equation}\label{eq:costJ1}
    J_{1,r} = \lambda_{\mathrm{tr}}\sum_t
    \big(\mathbf{z}'_{\mathrm{box},r}(t)-\mathbf{z}_{\mathrm{box}}^*(t)\big)^\top
    \Sigma_{\mathbf{z}_{\mathrm{box}}^*}(t)^{\dagger}
    \big(\mathbf{z}'_{\mathrm{box},r}(t)-\mathbf{z}_{\mathrm{box}}^*(t)\big),
\end{equation}
\begin{equation}\label{eq:costJ2}
    J_{2,r}=\sum_t\lVert\mathbf{w}_{\mathrm{env},r}(t)\rVert_2.
\end{equation}
where $\lambda_{\mathrm{tr}}>0$ scales the tracking penalty and $\dagger$ denotes the Moore-Penrose pseudoinverse, used to handle singular reference covariances. The first term \eqref{eq:costJ1} is a scaled Mahalanobis tracking cost. If covariance information is unavailable, the covariance is set to the identity, giving a scaled squared Euclidean tracking cost. The second term \eqref{eq:costJ2} penalizes object-environment contact through the sum of Euclidean norms of the contact wrench between the object and environment $\mathbf{w}_{\text{env},r}(t)$, which contains three force components and three moment components.

The rollouts are sorted by cost, with permutation $\pi$ satisfying
\begin{equation} \label{eq:sort}
    J_{\pi(1)} \le J_{\pi(2)} \le \cdots \le J_{\pi(R)} .
\end{equation}
The DMP parameters are updated using the lowest-cost rollout,
\begin{equation} \label{eq:updateTheta}
    \mathbf{\Theta}_d \leftarrow \mathbf{\Theta}_{d,\pi(1)},
    \quad d=1,\ldots,6 ,
\end{equation}
and the exploration covariance is updated using a variant inspired by the Cross-Entropy Method (CEM) \cite{stulp2012path}, based on the best $K_e\le R$ elite rollouts. The update below is centered on the selected lowest-cost parameter vector:
\begin{equation} \label{eq:CEM2}
\begin{aligned}
     \Sigma_{\mathbf{\Theta}_d} \leftarrow
    \frac{1}{K_e}\sum_{k=1}^{K_e}
    \big(\mathbf{\Theta}_{d,\pi(k)}-\mathbf{\Theta}_d\big)
    \big(\mathbf{\Theta}_{d,\pi(k)}-\mathbf{\Theta}_d\big)^\top, \\
    d=1,\ldots,6 .
\end{aligned}
\end{equation}
The procedure repeats until
\begin{equation} \label{eq:converge}
    \max_{i,j} \left(\Sigma_{\mathbf{\Theta}_d}\right)_{ij} < \epsilon_{\text{conv}},
    \quad d=1,\ldots,6 ,
\end{equation}
where $0<\epsilon_{\text{conv}}<c_d$. Upon convergence, the optimized DMP parameters define the refined object trajectory $\mathbf{z}_\text{box}^{\mathrm{ref}}(t)$, from which the end-effector reference trajectories $\mathbf{z}_L^\mathrm{ref}(t)$ and $\mathbf{z}_R^\mathrm{ref}(t)$ are obtained using the known grasp geometry.

\subsection{Phase II: Estimation of Object Parameters} \label{sec:methods_phase2}

\begin{figure}
    \centering
        \subfloat[]{\includegraphics[width=0.5\linewidth]{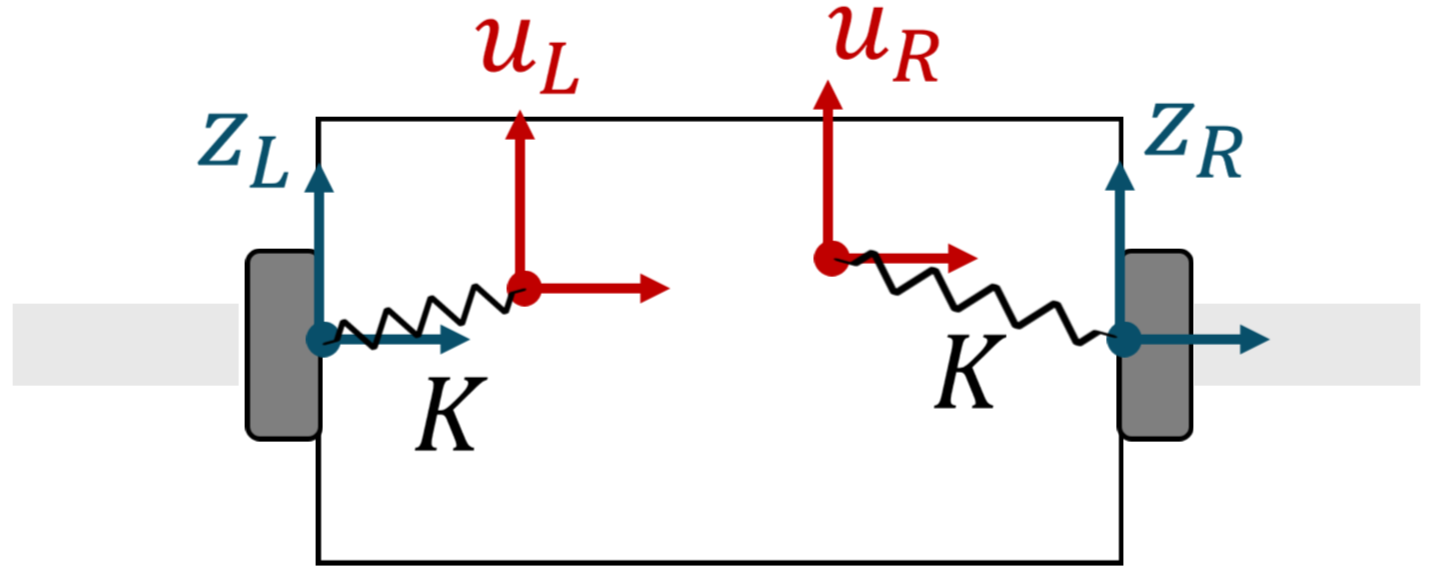}
        \label{fig:methods_phase2_box_impedance_control}}
    \hfil
        \subfloat[]{\includegraphics[width=0.45\linewidth]{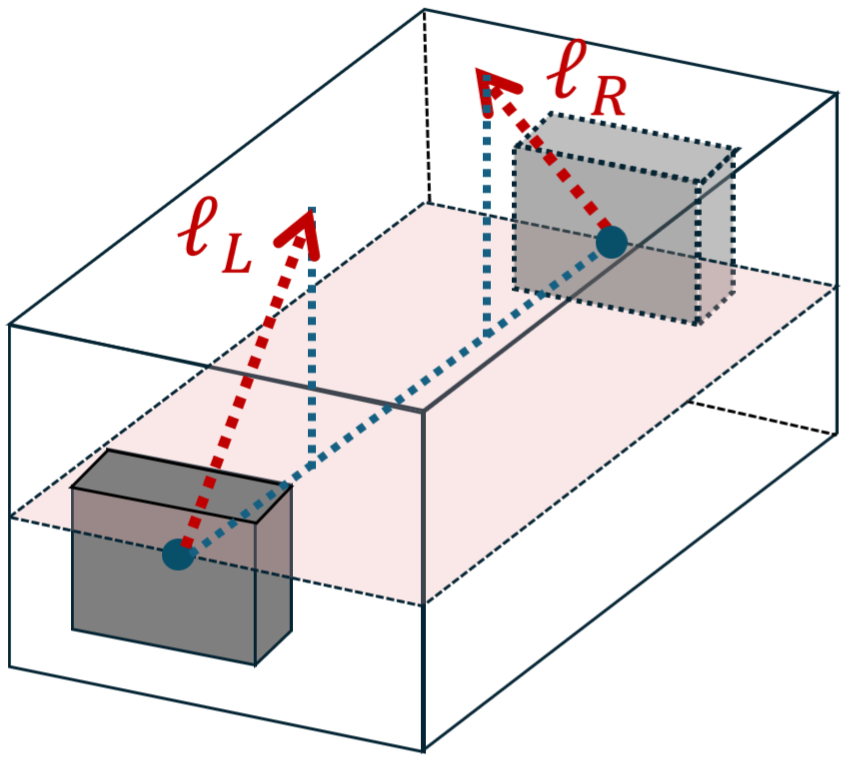}
        \label{fig:methods_phase2_lifting_direction_illustration}}
    \caption{Phase II overview. (a) Impedance-based interaction model during dual-arm lifting. Commanded task-space poses generate contact wrenches through virtual springs. (b) Lift initiation by ramping commanded end-effector poses along predefined directions.}
    \label{fig:Phase2}
\end{figure}

\begin{figure}
    \centering
    \includegraphics[width=\linewidth]{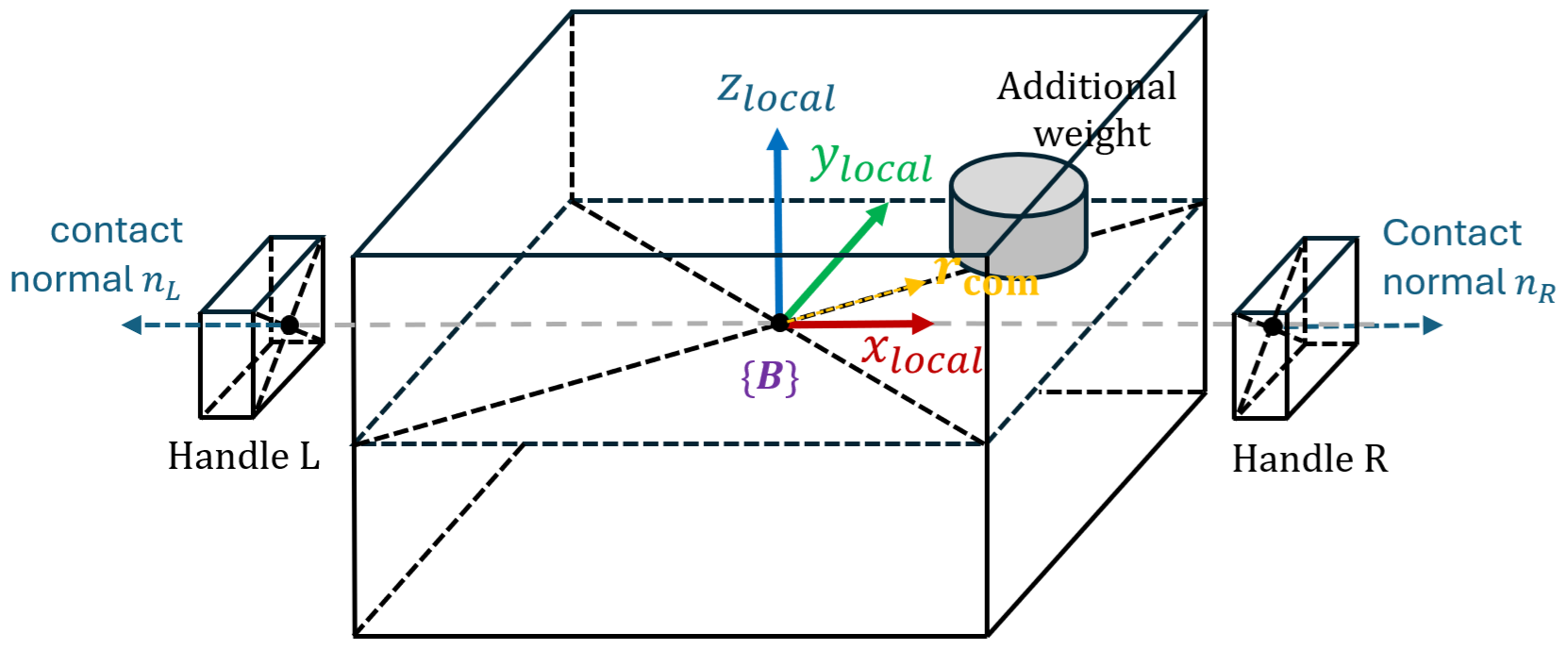}
    \caption{Object, left handle, and right handle coordinate frames. The box-fixed $\{B\}$ frame $\{x_{local}, y_{local}, z_{local}\}$ is defined at the geometric center, with $z_{local}$ aligned with the lifting direction. The center of mass (CoM) is expressed in this frame. The contact normal directions $\mathbf n_L$ and $\mathbf n_R$ at the left and right handles are indicated. An additional weight is used to vary the mass distribution.}
    \label{fig:box_frame_of_reference}
\end{figure}

Throughout the proposed framework, robot-object interaction is modeled through Cartesian impedance control, where the interaction wrench is proportional to the difference between the commanded and measured end-effector poses \cite{CAMPOLO2025116003}:
\begin{equation} \label{eq:impedance_control}
    \mathbf w = \mathbf K(\mathbf u-\mathbf z),
\end{equation}
where $\mathbf w \in \R^6$ is the task-space wrench, and $\mathbf u,\mathbf z\in\mathbb R^6$ are the commanded and measured end-effector poses, respectively. $\mathbf K\in\mathbb R^{6\times 6}$ is a positive definite stiffness matrix. Damping is omitted because it is handled by the low-level controller.

For each handle $i\in\{L,R\}$, the contact is represented by a point located at the center of the handle-box contact region. Its position relative to the box geometric center is denoted by $\mathbf r_i\in\mathbb R^3$. During Phase~II, the measured contact wrench is written as $\hat{\mathbf w}_i=[\hat{\mathbf f}_i^\top \; \hat{\boldsymbol{\tau}}_i^\top]^\top\in\mathbb R^6$, where $\hat{(\cdot)}$ denotes a measured or estimated quantity; unit vectors are written without hats. Here, $\hat{\mathbf f}_i$ is the measured force applied on the object and $\hat{\boldsymbol{\tau}}_i$ is the measured free contact moment about the contact point. Both are expressed in the object-fixed frame $\{B\}$ shown in \figref{fig:box_frame_of_reference}.

\subsubsection{Lift initiation and lift-off detection}
In Phase~II, the object mass and CoM are estimated after lift-off. To initiate lifting, the commanded end-effector poses are ramped along predefined unit lifting directions $\boldsymbol{\ell}_i$, whose orientation components are zero:
\begin{equation}
    \mathbf u_i(k)
    =
    \mathbf u_i(0) + \eta(k)\boldsymbol{\ell}_i,
    \qquad i\in\{L,R\},
\end{equation}
where $k$ is the discrete control step, $\mathbf u_i(0)$ is the initial commanded pose, and $\eta(k)$ is a scalar lift magnitude. The lift magnitude is initialized as $\eta(0)=0$ and increased monotonically:
\begin{equation}
    \eta(k+1)=\eta(k)+\Delta\eta ,
\end{equation}
where $\Delta\eta>0$ is the lift increment. This gradually increases the commanded lift until the object detaches from the ground. Lift-off is detected from end-effector feedback. Let $\mathbf p_i(k)\in\mathbb R^3$ denote the position component of the measured end-effector pose. The vertical displacement of each handle is
\begin{equation} \label{eq:handle_height}
    \Delta h_i(k)
    =
    \big(\mathbf p_i(k)-\mathbf p_i(0)\big)^\top
    \mathbf e_z,
    \qquad i\in\{L,R\}.
\end{equation}
Lift-off is declared when
\begin{equation} \label{eq:liftoff_condition}
    \min\{\Delta h_L(k),\Delta h_R(k)\}
    \ge h_{\mathrm{lift}},
\end{equation}
where $h_{\mathrm{lift}}$ is chosen conservatively to ensure clearance even under small object tilting.

\subsubsection{Mass estimation}
Once lift-off is achieved, the object is assumed to be in equilibrium, and the mass is estimated from force balance along the lifting direction. A set of $M$ post-lift-off measurements is collected, and the estimation is formulated as a least-squares problem. All quantities are expressed in the box-fixed frame $\{B\}$, where the contact locations and CoM are constant. The gravity vector expressed in $\{B\}$ is defined as
\begin{equation}
    \mathbf g = -g\,\mathbf e_z,
    \qquad g>0,
\end{equation}
where $g$ is the gravitational acceleration magnitude and $\mathbf e_z$ is the unit vector along $z_{\mathrm{local}}$ in the box frame.

For the $j$-th post-lift-off measurement, with $j=1,\ldots,M$, define the total measured contact force as
\begin{equation}
    \hat{\mathbf F}_j
    =
    \hat{\mathbf f}_{L,j}
    +
    \hat{\mathbf f}_{R,j}.
\end{equation}
Force equilibrium gives
\begin{equation}
    \hat{\mathbf F}_j
    +
    m\mathbf g
    \approx
    \mathbf 0,
\end{equation}
where $m$ is the object mass. Accounting for measurement noise and quasi-static modeling error, this can be written as
\begin{equation}
    \hat{\mathbf F}_j
    =
    -\mathbf g m
    +
    \boldsymbol{\epsilon}_{F,j}.
\end{equation}
where $\boldsymbol{\epsilon}_{F,j}$ captures measurement noise and quasi-static modeling error. Stacking all measurements gives the linear regression model:
\begin{equation}
    \mathbf y_F
    =
    \boldsymbol{\Phi}_F m
    +
    \boldsymbol{\epsilon}_F,
\end{equation}
where
\begin{equation}
    \mathbf y_F
    =
    \begin{bmatrix}
        \hat{\mathbf F}_1 \\
        \vdots \\
        \hat{\mathbf F}_{M}
    \end{bmatrix}
    \in \mathbb R^{3M},
    \qquad
    \boldsymbol{\Phi}_F
    =
    \begin{bmatrix}
        -\mathbf g \\
        \vdots \\
        -\mathbf g
    \end{bmatrix}
    \in \mathbb R^{3M\times 1}.
\end{equation}
The mass estimate is obtained as
\begin{equation}
    \hat m
    =
    \boldsymbol{\Phi}_F^\dagger
    \mathbf y_F ,
\end{equation}
where $(\cdot)^\dagger$ denotes the Moore-Penrose pseudoinverse.

\subsubsection{CoM estimation}
The CoM is expressed in the object-fixed frame as
\begin{equation}
    \mathbf r_{\mathrm{com}}
    =
    \begin{bmatrix}
        r_x & r_y & r_z
    \end{bmatrix}^{\top}.
\end{equation}
For the $j$-th measurement, define the resultant measured contact moment about the box geometric center as
\begin{equation}
    \hat{\mathbf s}_j
    =
    \sum_{i\in\{L,R\}}
    \left(
        \mathbf r_i \times \hat{\mathbf f}_{i,j}
        +
        \hat{\boldsymbol{\tau}}_{i,j}
    \right),
\end{equation}
where $\mathbf r_i$ is the vector from the box geometric center to contact $i$. Moment equilibrium gives
\begin{equation}
    \hat{\mathbf s}_j
    +
    \mathbf r_{\mathrm{com}}
    \times
    \left(
        \hat m\mathbf g
    \right)
    \approx
    \mathbf 0.
\end{equation}
Rewriting the cross product in skew-symmetric matrix form and accounting for measurement noise and quasi-static modeling error gives
\begin{equation}
    \hat{\mathbf s}_j
    =
    [\hat m\mathbf g]_{\times}
    \mathbf r_{\mathrm{com}}
    +
    \boldsymbol{\epsilon}_{r,j}.
\end{equation}
Stacking all measurements gives the linear regression model:
\begin{equation}
    \mathbf y_r
    =
    \boldsymbol{\Phi}_r
    \mathbf r_{\mathrm{com}}
    +
    \boldsymbol{\epsilon}_r,
\end{equation}
where
\begin{equation}
    \mathbf y_r
    =
    \begin{bmatrix}
        \hat{\mathbf s}_1 \\
        \vdots \\
        \hat{\mathbf s}_{M}
    \end{bmatrix}
    \in \mathbb R^{3M},
    \qquad
    \boldsymbol{\Phi}_r
    =
    \begin{bmatrix}
        [\hat m\mathbf g]_{\times} \\
        \vdots \\
        [\hat m\mathbf g]_{\times}
    \end{bmatrix}
    \in \mathbb R^{3M\times 3}.
\end{equation}
The CoM estimate is then
\begin{equation}
    \hat{\mathbf r}_{\mathrm{com}}
    =
    \boldsymbol{\Phi}_r^\dagger
    \mathbf y_r .
\end{equation}
The rank of $\boldsymbol{\Phi}_r$ determines which CoM components are observable. In the present lifting setting, gravity acts along $z_{\mathrm{local}}$, so only the in-plane CoM coordinates affect the lifting moment equilibrium. The component $r_z$ does not contribute to the gravitational moment during upright lifting. Therefore, the estimated in-plane coordinates $\hat r_x$ and $\hat r_y$ are used in Phase~III for contact wrench optimization.

\subsection{Phase III: Contact Force and Moment Optimization via Second-Order Cone Programming} \label{sec:methods_phase3}

In Phase~III, the objective is to compute optimal contact wrenches at the two robot end-effectors that support the object against gravity while respecting frictional contact constraints. The formulation is based on the quasi-static assumption and uses the online estimated object mass and CoM obtained in Phase~II.

\subsubsection{Contact model}

Let $\mathbf n_i\in\mathbb R^3$ denote the outward unit normal at contact $i\in\{L,R\}$, see \figref{fig:box_frame_of_reference}, $\mathbf f_i\in\mathbb R^3$ denote the contact force and $\boldsymbol{\tau}_i\in\mathbb R^3$ denote the contact moment about the contact point, both expressed in the object-fixed frame $\{B\}$. The orthogonal projector onto the tangent plane is defined as
\begin{equation}
    \boldsymbol{\Pi}_i
    =
    \mathbf I_3-\mathbf n_i\mathbf n_i^\top .
\end{equation}

The contact force $\mathbf f_i$ is decomposed into a signed normal component $f_{n,i}$ and a tangential component $\mathbf f_{t,i}$, defined as
\begin{equation}
    f_{n,i}
    =
    \mathbf n_i^\top \mathbf f_i,
    \qquad
    \mathbf f_{t,i}
    =
    \boldsymbol{\Pi}_i\mathbf f_i,
    \label{eq:f_ni_andf_ti}
\end{equation}
where compression corresponds to $f_{n,i}\le 0$. 

Similarly, the contact moment $\boldsymbol{\tau}_i$ is decomposed into its torsional and tangential components as
\begin{equation}
    \tau_{n,i}
    =
    \mathbf n_i^\top \boldsymbol{\tau}_i,
    \qquad
    \boldsymbol{\tau}_{t,i}
    =
    \boldsymbol{\Pi}_i\boldsymbol{\tau}_i .
    \label{eq:tau_ni_andtau_ti}
\end{equation}

Since rotational balance due to CoM offsets can be achieved by the contact forces together with the torsional moment about the contact normal, the contact model does not require bending moments about the tangent directions. Therefore, the following constraint is imposed
\begin{equation}
    \boldsymbol{\tau}_{t,i}
    =
    \boldsymbol{\Pi}_i\boldsymbol{\tau}_i
    =
    \mathbf 0.
    \label{eq:no_bending_moment}
\end{equation}

\subsubsection{Equilibrium constraints}

Under quasi-static conditions, the desired contact wrenches must balance the estimated gravitational wrench. Force and moment equilibrium about the box geometric center require
\begin{align}
    \sum_{i\in\{L,R\}} \mathbf f_i
    +
    \hat m \mathbf g
    &=
    \mathbf 0,
    \label{eq:phase3_force_balance}
    \\
    \sum_{i\in\{L,R\}}
    \left(
        \mathbf r_i \times \mathbf f_i
        +
        \boldsymbol{\tau}_i
    \right)
    +
    \hat{\mathbf r}_{\mathrm{com}}
    \times
    \hat m \mathbf g
    &=
    \mathbf 0,
    \label{eq:phase3_moment_balance}
\end{align}
where $\mathbf r_i$ is the vector from the box geometric center to contact $i$.

\subsubsection{Wrench redundancy and nullspace interpretation}

The equilibrium constraints define the object-level wrench required for stable lifting, but they do not uniquely determine the individual contact wrenches. For each contact $i\in\{L,R\}$, define the contact wrench as ${\mathbf w}_i=[{\mathbf f}_i^\top \; {\boldsymbol{\tau}}_i^\top]^\top\in\mathbb R^6$. To make the wrench redundancy explicit, define the stacked contact wrench
\begin{equation}
    \mathbf x
        =
    \begin{bmatrix}
        \mathbf w_L^\top &
        \mathbf w_R^\top
    \end{bmatrix}^{\top}
    =
    \begin{bmatrix}
        \mathbf f_L^\top &
        \boldsymbol{\tau}_L^\top &
        \mathbf f_R^\top &
        \boldsymbol{\tau}_R^\top
    \end{bmatrix}^{\top}
    \in \mathbb R^{12}.
    \label{eq:stacked_contact_wrench}
\end{equation}
The net wrench applied to the object about the box geometric center can be written as
\begin{equation}
    \mathbf w_{\mathrm{obj}}
    =
    \mathbf G\mathbf x,
    \label{eq:grasp_map}
\end{equation}
where the grasp map is
\begin{equation}
    \mathbf G
    =
    \begin{bmatrix}
        \mathbf I_3 & \mathbf 0_3 & \mathbf I_3 & \mathbf 0_3 \\
        [\mathbf r_L]_\times & \mathbf I_3 &
        [\mathbf r_R]_\times & \mathbf I_3
    \end{bmatrix}
    \in \mathbb R^{6\times 12}.
    \label{eq:grasp_map_matrix}
\end{equation}
Here, $[\mathbf r_i]_\times\mathbf f_i=\mathbf r_i\times\mathbf f_i$. The equilibrium constraints \eqref{eq:phase3_force_balance}-\eqref{eq:phase3_moment_balance} can therefore be written compactly as
\begin{equation}
    \mathbf G\mathbf x
    =
    -
    \begin{bmatrix}
        \hat m\mathbf g\\
        \hat{\mathbf r}_{\mathrm{com}}\times \hat m\mathbf g
    \end{bmatrix}.
    \label{eq:compact_equilibrium}
\end{equation}
Since $\mathbf G$ maps a 12-dimensional contact wrench vector to a 6-dimensional object wrench, the contact wrench distribution is generally redundant at the equilibrium level. This means that there can be nonzero contact-wrench variations $\mathbf x_{\mathrm{null}}$ that satisfy
\begin{equation}
    \mathbf G\mathbf x_{\mathrm{null}}=\mathbf 0.
    \label{eq:nullspace_condition}
\end{equation}
Such variations lie in the kernel, or nullspace, of $\mathbf G$, and therefore do not change the net wrench applied to the box. Physically, $\mathbf x_{\mathrm{null}}$ represents an internal wrench: it changes how forces and moments are distributed between the two contacts without changing the object-level equilibrium. Next, the friction constraints and minimum-effort objective are introduced to select one physically feasible wrench distribution from this redundant set.

\subsubsection{Friction constraints}
Slip at each contact is prevented using a friction limit surface \cite{howe1996practical}, which captures the coupling between tangential force and torsional moment. For contact $i\in\{L,R\}$, the limit surface is modeled using the ellipsoidal approximation \cite{xydas1999modeling} as
\begin{equation} \label{eq:limitsurface_ellipsoid}
    \left(
    \frac{\|\mathbf{f}_{t,i}\|_2}{f_{t,i}^{\max}}
    \right)^2
    +
    \left(
    \frac{\tau_{n,i}}{\tau_{n,i}^{\max}}
    \right)^2
    \le (1-r_s)^2.
\end{equation}
Here, $r_s\in[0,1)$ defines a safety margin through the Pontryagin difference of the normalized friction region and a perturbation ball of radius $r_s$. The contracted surface bounds a region of radius $(1-r_s)$, accommodating normalized tangential-force and torsional-moment perturbations up to $r_s$ at fixed normal force and friction parameters. See ~\ref{app:safety_margin_rs} for the derivation.

The size of the friction limit surface depends on the contact normal force, the friction coefficient, and the contact geometry. Since compression corresponds to $f_{n,i}\leq0$, the compressive normal-force magnitude is $-f_{n,i}$. The maximum admissible tangential force is therefore given by Coulomb friction as
\begin{equation}
    f_{t,i}^{\max}
    =
    \mu(-f_{n,i}),
    \label{eq:f_tan_max}
\end{equation}
while the maximum admissible torsional moment about the contact normal is
\begin{equation}
    \tau_{n,i}^{\max}
    =
    \mu(-f_{n,i})R_{\mathrm{eff},i},
    \label{eq:tau_n_max}
\end{equation}
where $\mu$ is the friction coefficient and $R_{\mathrm{eff},i}$ is an effective contact radius determined by the contact geometry and pressure distribution \cite{howe1996practical,xydas1999modeling}. The derivation of $R_{\mathrm{eff},i}$ is given in ~\ref{app:torsional_friction}. 

For $f_{n,i}<0$, substituting \eqref{eq:f_tan_max} and \eqref{eq:tau_n_max} into \eqref{eq:limitsurface_ellipsoid} gives
\begin{equation}
    \frac{\|\mathbf f_{t,i}\|_2^2}
    {\mu^2(-f_{n,i})^2}
    +
    \frac{\tau_{n,i}^2}
    {\mu^2(-f_{n,i})^2R_{\mathrm{eff},i}^2}
    \le
    (1-r_s)^2 .
    \label{eq:limitsurface_substituted}
\end{equation}
Rearranging \eqref{eq:limitsurface_substituted} yields
\begin{equation}
    \|\mathbf f_{t,i}\|_2^2
    +
    \left(
    \frac{\tau_{n,i}}{R_{\mathrm{eff},i}}
    \right)^2
    \le
    \left[
    (1-r_s)\mu(-f_{n,i})
    \right]^2 .
    \label{eq:limitsurface_rearranged}
\end{equation}
Equivalently, the friction limit can be written in second-order-cone (SOC) form \cite{boyd2004convex} as
\begin{equation}
    \left\|
    \begin{bmatrix}
        \mathbf f_{t,i}\\[1mm]
        \tau_{n,i}/R_{\mathrm{eff},i}
    \end{bmatrix}
    \right\|_2
    \le
    (1-r_s)\mu(-f_{n,i}).
    \label{eq:friction_soc}
\end{equation}
Using \eqref{eq:f_ni_andf_ti} and \eqref{eq:tau_ni_andtau_ti}, \eqref{eq:friction_soc} can be expressed directly in terms of the contact-wrench components as
\begin{equation}
    \left\|
    \begin{bmatrix}
        \boldsymbol{\Pi}_i\mathbf f_i\\[1mm]
        \dfrac{\mathbf n_i^\top\boldsymbol{\tau}_i}
        {R_{\mathrm{eff},i}}
    \end{bmatrix}
    \right\|_2
    \le
    (1-r_s)\mu
    \left(-\mathbf n_i^\top\mathbf f_i\right).
    \label{eq:friction_soc_explicit}
\end{equation}
This has the standard SOC structure \cite{boyd2004convex}: the left-hand side is the Euclidean norm of a linear function of the contact wrench, while the right-hand side is linear in the compressive normal force and is nonnegative under the condition $f_{n,i}\leq0$. Thus, \eqref{eq:friction_soc_explicit} provides a convex representation of the coupled tangential-force and torsional-moment friction limit, including its closed extension at zero normal force.

\subsubsection{Optimization problem}
Using the stacked contact wrench $\mathbf x$ defined in \eqref{eq:stacked_contact_wrench}, the contact wrenches are selected by minimizing a weighted wrench effort. For each contact wrench $\mathbf w_i$, define the weighting matrix
\begin{equation}
    \mathbf Q_c
    =
    \operatorname{diag}
    \left(
    1,1,1,
    \frac{1}{l_c^2},
    \frac{1}{l_c^2},
    \frac{1}{l_c^2}
    \right),
    \label{eq:contact_weight_matrix}
\end{equation}
where $l_c>0$ is a characteristic contact length chosen from the contact patch size to account for the different units of force and moment in the weighted wrench cost \cite{lynch2017modern}. Its selection from the contact patch dimensions is a design choice in the present framework. The stacked weighting matrix is then
\begin{equation}
    \mathbf Q
    =
    \operatorname{blkdiag}(\mathbf Q_c,\mathbf Q_c).
    \label{eq:stacked_weight_matrix}
\end{equation}
The contact wrenches are selected by minimizing the weighted wrench effort:
\begin{equation}
    \min_{\mathbf x} \ \mathbf x^\top \mathbf Q \mathbf x
    =
    \min_{\mathbf x}
    \left(
        \mathbf w_L^\top \mathbf Q_c \mathbf w_L
        +
        \mathbf w_R^\top \mathbf Q_c \mathbf w_R
    \right).
    \label{eq:minx}
\end{equation}
Since $\mathbf Q$ is positive definite, minimizing the squared weighted effort $\mathbf x^\top\mathbf Q\mathbf x$ has the same minimizer as minimizing $\|\mathbf Q^{1/2}\mathbf x\|_2$. Introducing an auxiliary variable $\gamma\in\mathbb R$, the optimization problem is written in epigraph form as
\begin{equation}
\begin{aligned}
\min_{\mathbf x,\gamma} \quad & \gamma \\
\text{s.t.} \quad
& \mathbf n_i^\top \mathbf f_i \le 0,
  \quad i\in\{L,R\},
    \quad \text{(compression)}\\
& \boldsymbol{\Pi}_i\boldsymbol{\tau}_i = \mathbf 0,
  \quad i\in\{L,R\},
  \quad \text{(no bending moment)} \\
& \text{equilibrium constraints }
  \eqref{eq:compact_equilibrium}, \\
& \text{friction limit constraints } \eqref{eq:friction_soc_explicit},
  \qquad i\in\{L,R\}, \\
& \left\|\mathbf Q^{1/2}\mathbf x\right\|_2 \le \gamma ,
\end{aligned}
\label{eq:phase3_socp_reduced}
\end{equation}
where $\mathbf Q^{1/2}$ is the square-root weighting matrix. The equilibrium constraints in \eqref{eq:compact_equilibrium} and no-bending-moment constraints are affine equalities, while the compression constraints are affine inequalities. The friction constraints in \eqref{eq:friction_soc_explicit} and the objective epigraph constraint $\|\mathbf Q^{1/2}\mathbf x\|_2\le \gamma$ are second-order cone constraints. Therefore, \eqref{eq:phase3_socp_reduced} is a convex second-order cone program (SOCP), which can be solved efficiently using standard convex optimization methods \cite{boyd2004convex}. 

The resulting optimal contact wrench for each handle is denoted by $\mathbf w_i^{\mathrm{des}} = \bigl[(\mathbf f_i^{\mathrm{des}})^\top \; (\boldsymbol{\tau}_i^{\mathrm{des}})^\top\bigr]^\top \in \mathbb R^6$, $i\in\{L,R\}$, where $\mathbf f_i^{\mathrm{des}}$ and $\boldsymbol{\tau}_i^{\mathrm{des}}$ are the desired contact force and moment, respectively. These wrenches are passed to the execution layer for quasi-static tracking of the reference trajectory obtained in Phase~I.

\subsection{Execution} \label{sec:execution_layer}

Phase~III provides the desired contact wrenches $\mathbf{w}_{i}^{\mathrm{des}}$, $i\in\{L,R\}$. These wrenches are executed along the end-effector references $\mathbf{z}_i^{\mathrm{ref}}(t)$, which are obtained from the refined object trajectory $\mathbf{z}_\mathrm{box}^{\mathrm{ref}}(t)$ from Phase~I using the known grasp geometry. Using the Cartesian impedance relation introduced in \figref{fig:methods_phase2_box_impedance_control}, the desired wrench is realized by biasing the commanded pose relative to the reference trajectory. The nominal command is defined as
\begin{equation}
    \mathbf{u}_{i,0}(t)
    =
    \mathbf{z}_i^{\mathrm{ref}}(t)
    +
    \mathbf{K}^{-1}\mathbf{w}_i^{\mathrm{des}},
    \label{eq:nominal_command}
\end{equation}
where $\mathbf K\ \in \R^{6 \times 6}$ is the Cartesian stiffness matrix. Thus, both the desired contact force and contact moment obtained in Phase~III are converted into corresponding translational and rotational pose offsets through the impedance model. 

In practice, the realized contact wrench can differ from its desired value due to actuator dynamics, sensor noise, and environmental uncertainty \cite{turlapati2024identification}. Therefore, an additional force-feedback correction is used during execution. Using the measured contact force $\hat{\mathbf f}_i(t)$, expressed with the same force-on-object convention and in the same frame as the desired force, the force tracking error is defined as
\begin{equation}
    \mathbf e_{f,i}(t)
    =
    \mathbf f_i^{\mathrm{des}}
    -
    \hat{\mathbf f}_i(t),
    \qquad i\in\{L,R\}.
    \label{eq:force_feedback_error}
\end{equation}
A proportional-integral-derivative (PID) controller maps the force tracking error to a translational correction of the commanded end-effector pose:
\begin{equation}
    \Delta\mathbf p_i(t)
    =
    \mathbf K_P\mathbf e_{f,i}(t)
    +
    \mathbf K_I
    \int_0^t \mathbf e_{f,i}(s)\,ds
    +
    \mathbf K_D\dot{\mathbf e}_{f,i}(t),
    \label{eq:force_pid_correction}
\end{equation}
where $\mathbf K_P,\mathbf K_I,\mathbf K_D\in\mathbb R^{3\times3}$ are the proportional, integral, and derivative gain matrices, respectively. The corrective Cartesian pose increment is defined as
\begin{equation}
    \Delta\mathbf u_i(t)
    =
    \begin{bmatrix}
        \Delta\mathbf p_i(t)\\
        \mathbf 0_3
    \end{bmatrix}
    \in\mathbb R^6.
    \label{eq:pose_feedback_correction}
\end{equation}
The final Cartesian command is therefore
\begin{equation}
    \mathbf u_i(t)
    =
    \mathbf u_{i,0}(t)
    +
    \Delta\mathbf u_i(t),
    \label{eq:final_command}
\end{equation}
where $\mathbf u_{i,0}(t)$ is the nominal impedance command in \eqref{eq:nominal_command}. Thus, the desired contact wrench is determined by the Phase~III optimization, while the feedback controller compensates for discrepancies between the desired and measured contact forces during physical execution.

\section{Experiments and Results}
\label{sec:experiment_and_result}

\subsection{Experiment Setup}
\label{sec:experiment_setup}

\begin{figure}[!t]
    \centering
        \subfloat[]{\includegraphics[width=0.5\linewidth]{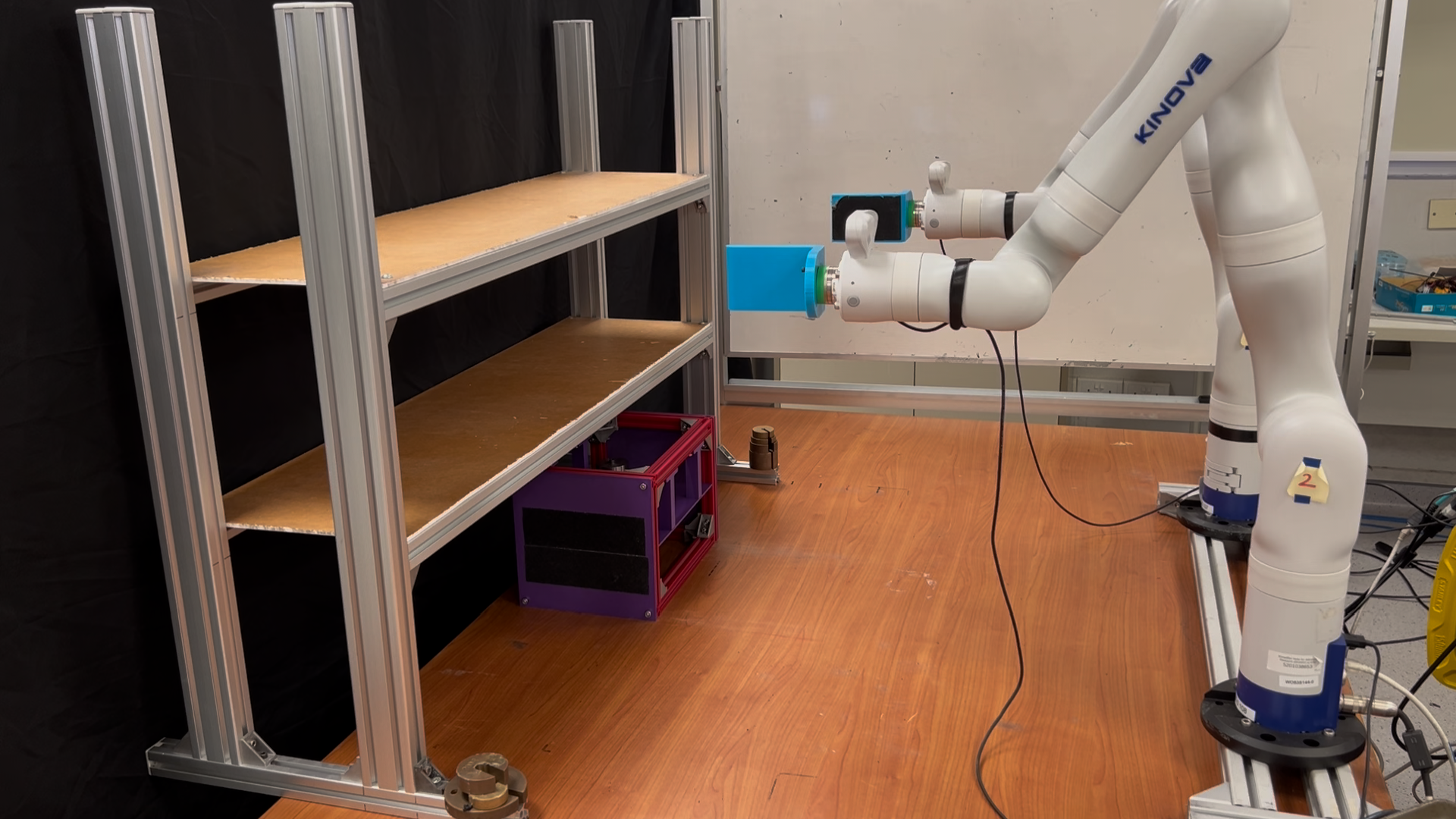}
        \label{fig:experiment_setup}}
    \hfil
        \subfloat[]{\includegraphics[width=0.2\linewidth]{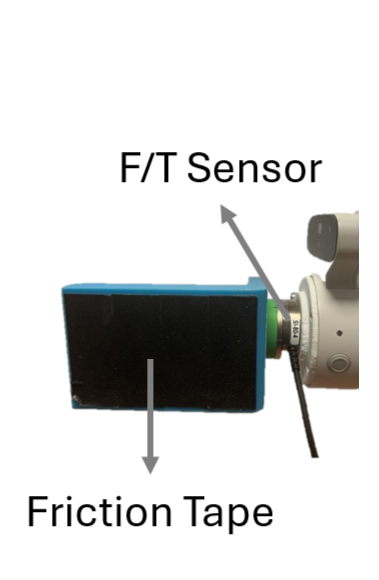}
        \label{fig:experiment_setup_ftsensor}}
    \hfil
        \subfloat[]{\includegraphics[width=0.25\linewidth]{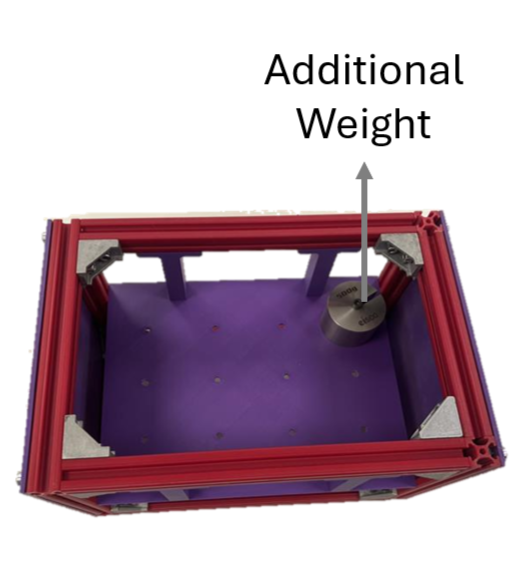}
        \label{fig:experiment_setup_box}}
    \hfil
        \subfloat[]{\includegraphics[width=0.7\linewidth] {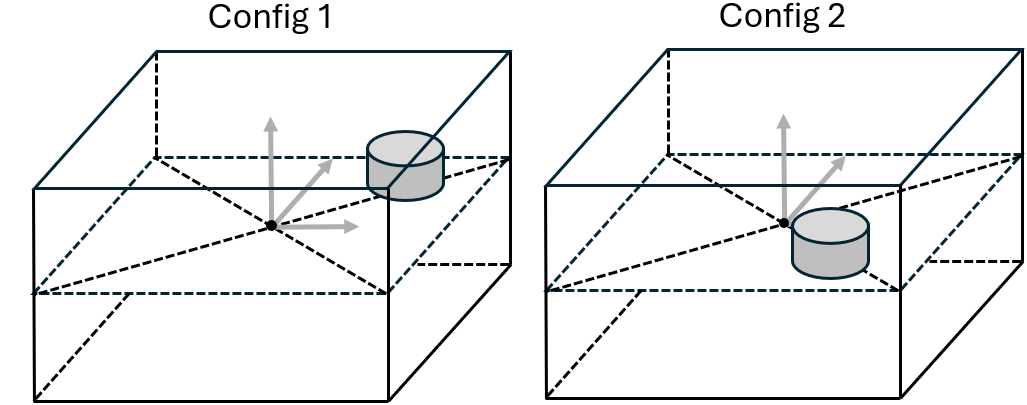}
        \label{fog:experiment_setup_two_config}}
    \hfil
    \caption{Experimental setup for dual-arm box manipulation. (a) Dual-arm robotic system interacting with the box in a shelf environment. (b) Close-up of the end-effector equipped with an ATI Mini40 force/torque (F/T) sensor and a friction-enhanced contact surface. (c) The test box with an additional weight used to vary the center-of-mass (CoM) location. (d) Illustration of the two CoM configurations used in the experiment.}
    \label{fig:experiment_setup_all}
\end{figure}

The experimental setup is shown in \figref{fig:experiment_setup_all}. The system consists of two 7-DoF Kinova robotic arms, each equipped with an ATI Mini40 force/torque (F/T) sensor at the wrist. A custom handle with a friction-enhanced contact surface (coefficient of friction 0.4) is attached to each sensor. The handle contact patch has dimensions of $100\,\text{mm} \times 70\,\text{mm}$. The manipulated object is a rectangular box with a base mass of 1.7\,kg. An additional mass of 0.5\,kg is placed inside the box to vary the CoM location.

The Cartesian impedance controller uses the stiffness matrix $\mathbf K=\operatorname{blkdiag}(1000\,\mathbf I_3,\, 10\,\mathbf I_3)$, corresponding to translational stiffness of $1000\,\mathrm{N/m}$ and rotational stiffness of $10\,\mathrm{Nm/rad}$, respectively. For the force-feedback controller in \eqref{eq:force_pid_correction}, the proportional gain is selected as the inverse of the translational stiffness, giving $\mathbf K_P=10^{-3}\mathbf I_3\,\mathrm{m/N}$. The integral gain is defined using an integral time of $T_{\mathrm I}=10\,\mathrm{s}$, resulting in $\mathbf K_I=10^{-4}\mathbf I_3\,\mathrm{m/(N\,s)}$. The derivative gain $\mathbf{K}_D$ is set to zero, as damping is already provided by the robot's low-level Cartesian impedance controller. The translational feedback correction is applied primarily in the squeezing and vertical load-support directions.

Two CoM configurations are considered in the experiments. In Configuration~1, the additional mass is placed with an offset in the positive $x_{local}$ and $y_{local}$ directions. In Configuration~2, the additional mass is placed with an offset in the positive $x_{local}$ and negative $y_{local}$ directions. Each configuration is repeated three times, resulting in three independent runs per configuration.

\subsection{Phase I Results} (Trajectory Refinement for Collision Avoidance)
\label{sec:phase1result}

\begin{figure}[t]
    \centering
        \subfloat[]{
        \includegraphics[width=\linewidth]{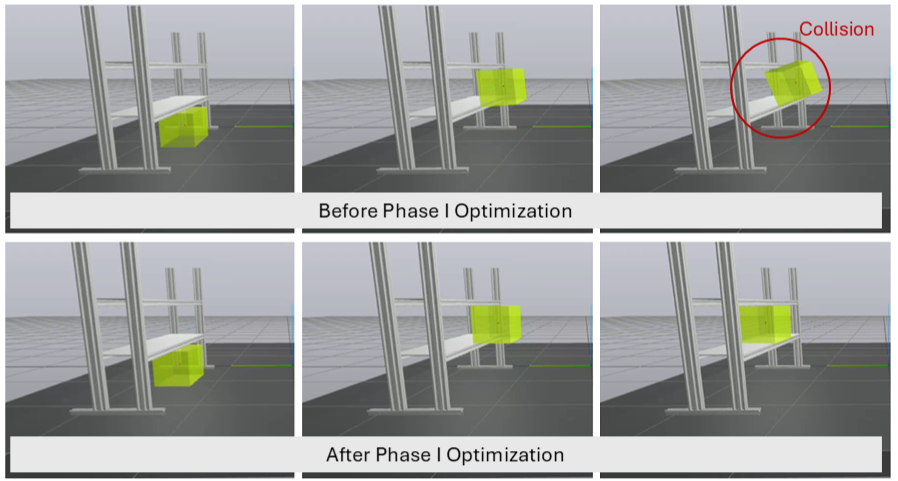}
        \label{fig:result_phase_1_picture}
        }
    \vspace{0.5em}
        \subfloat[]{
        \includegraphics[width=\linewidth]{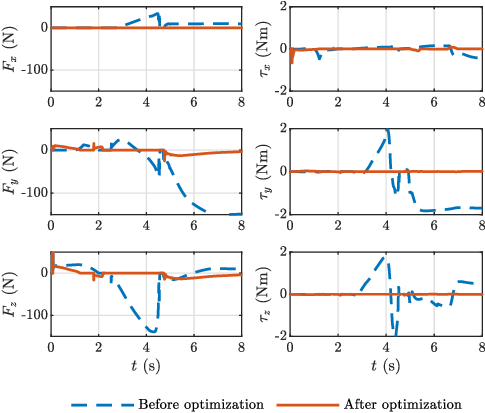}
        \label{fig:result_phase1_plot}
        } 
    \caption{Phase I optimization results. (a) Object motion before (top) and after (bottom) Phase I trajectory refinement, evaluated in the Drake physics engine. (b) Corresponding environment contact wrench before and after Phase I refinement. After optimization, most force and torque components remain close to zero, indicating successful collision avoidance.}
    \label{fig:result_phase1_combined}
\end{figure}

Phase~I refinement is demonstrated for Configuration~1, where the nominal trajectory produces object-environment contact during extraction. Configuration~2 does not require refinement because its nominal trajectory is already collision-free in the simulated environment. 

For Configuration~1, the initial object trajectory $\mathbf{z}_\text{box}^*(t)$ was generated using \cite{lee2025generalizingrobottrajectoriessinglecontext}. The trajectory refinement was performed in Drake \cite{drake} with $R=50$ rollouts per iteration, $\alpha=0.2$ in \eqref{eq:costJ} to balance trajectory tracking and contact avoidance, $\lambda_{\mathrm{tr}}=0.1$ in \eqref{eq:costJ1}, $K_e=5$ elite samples for the update in \eqref{eq:CEM2}, and $\epsilon_{\text{conv}}=10^{-2}$ in \eqref{eq:converge} as the convergence threshold. In this work, the undesired collision is primarily associated with object motion along the $y_{\mathrm{box}}$ and $z_{\mathrm{box}}$ directions. Thus, trajectory exploration is restricted to these two position components by setting $c_d=1000$. For the remaining position and orientation components, $c_d=0$, so their reference trajectories remain unchanged.

\figref{fig:result_phase1_combined}(a) shows the object motion before and after trajectory refinement. The initial trajectory produces collision with the environment, whereas the refined trajectory avoids the unintended contact while remaining close to the original motion. The environment contact wrench in \figref{fig:result_phase1_combined}(b) represents the wrench exerted by the environment on the object, where the environment may correspond to either the ground or the shelf depending on the stage of the motion. Before refinement, the nominal trajectory produces large contact force and torque components, indicating undesired object-environment collision during extraction. These large components correspond to contact between the box and the shelf. After refinement, most force and torque components remain close to zero, confirming that the unintended contact is reduced. The nonzero $f_y$ and $f_z$ components near the end of the motion correspond to the intended box-shelf interaction during placement, where $f_y$ represents friction between the box and the shelf surface and $f_z$ represents the normal support from the shelf. The refined trajectory is then used as the reference object motion in the execution experiments in Section~\ref{sec:execution_result}.

To evaluate the computation time of Phase~I, the optimization was repeated using ten independent random seeds for Configuration~1, with the same reference trajectory and optimization settings. The covariance convergence threshold was reached for all ten seeds, requiring $21.0 \pm 3.6$ iterations and $74.6 \pm 9.6\,\mathrm{s}$ of optimization time, reported as mean $\pm$ sample standard deviation across seeds. Including initialization, parallel-worker startup, and the initial and final simulations, the total elapsed time was $89.7 \pm 9.8\,\mathrm{s}$. These measurements were obtained using MATLAB R2026a and Drake on an Intel Core Ultra 9 285K processor with 24 parallel workers.

\subsection{Phase II Results} (Estimation of Object Parameters)
\label{sec:phase2result}

\begin{table}[!t]
\caption{Mass estimation results for two configurations.}
\label{tab:mass_estimation}
\centering
\footnotesize
\begin{tabular}{|c|c|c|c|c|}
\hline
\textbf{Config} & \textbf{Run} & \textbf{Experiment} & \textbf{Mass (kg)} & \textbf{Error (\%)} \\
\hline\hline
\multirow{7}{*}{1}
& -    & GT & 2.20   & - \\
\cline{2-5}
& \multirow{2}{*}{Run 1} & Sim & 2.2016 & 0.07 \\
&                        & Real & 2.3073 & 4.88 \\
\cline{2-5}
& \multirow{2}{*}{Run 2} & Sim & 2.1998 & 0.01 \\
&                        & Real & 2.2989 & 4.50 \\
\cline{2-5}
& \multirow{2}{*}{Run 3} & Sim & 2.1993 & 0.03 \\
&                        & Real & 2.3292 & 5.87 \\
\hline
\multirow{7}{*}{2}
& -    & GT & 2.20   & - \\
\cline{2-5}
& \multirow{2}{*}{Run 1} & Sim & 2.2016 & 0.07 \\
&                        & Real & 2.2839 & 3.81 \\
\cline{2-5}
& \multirow{2}{*}{Run 2} & Sim & 2.1990 & 0.05 \\
&                        & Real & 2.3528 & 6.04 \\
\cline{2-5}
& \multirow{2}{*}{Run 3} & Sim & 2.2003 & 0.01 \\
&                        & Real & 2.3918 & 8.72 \\
\hline
\end{tabular}
\end{table}

\begin{table}[!t]
\caption{Center-of-mass (CoM) estimation results for two configurations.}
\label{tab:com_estimation}
\centering
\footnotesize
\begin{tabular}{|c|c|c|c|c|}
\hline
\textbf{Config} & \textbf{Run} & \textbf{Experiment} & \textbf{CoM $(r_x, r_y)$ (mm)} & \textbf{Error (\%)} \\
\hline\hline
\multirow{7}{*}{1}
& -    & GT & $(20.50,\; 11.40)$ & - \\
\cline{2-5}
& \multirow{2}{*}{Run 1} & Sim & $(20.51,\; 11.44)$ & 0.01 \\
&                        & Real & $(8.58,\; 14.19)$ & 3.63 \\
\cline{2-5}
& \multirow{2}{*}{Run 2} & Sim & $(20.53,\; 11.40)$ & 0.01 \\
&                        & Real & $(7.65,\; 13.87)$ & 3.90 \\
\cline{2-5}
& \multirow{2}{*}{Run 3} & Sim & $(20.47,\; 11.39)$ & 0.03 \\
&                        & Real & $(8.21,\; 14.33)$ & 3.76 \\
\hline
\multirow{7}{*}{2}
& -    & GT & $(6.80,\; -11.40)$ & - \\
\cline{2-5}
& \multirow{2}{*}{Run 1} & Sim & $(6.79,\; -11.42)$ & 0.02 \\
&                        & Real & $(5.31,\; -16.01)$ & 3.34 \\
\cline{2-5}
& \multirow{2}{*}{Run 2} & Sim & $(6.85,\; -11.48)$ & 0.03 \\
&                        & Real & $(1.65,\; -16.48)$ & 4.50 \\
\cline{2-5}
& \multirow{2}{*}{Run 3} & Sim & $(6.83,\; -11.47)$ & 0.03 \\
&                        & Real & $(3.00,\; -18.79)$ & 4.49 \\
\hline
\end{tabular}
\end{table}

\begin{figure}
    \centering
    \includegraphics[width=\linewidth]{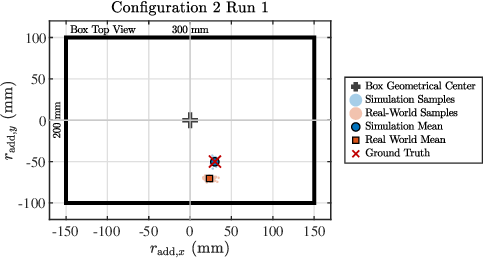}
    \caption{Phase II CoM estimation results visualized as the inferred additional-mass location for Configuration~2 (Run~1). The additional-mass location is inferred from the estimated CoM of the combined box-additional mass system using the known box mass and additional mass. Simulation estimates closely align with the ground truth, while real-world experiment estimates show a consistent offset but remain tightly clustered, indicating low variability. The observed deviations in the real-world are on the order of a few percent relative to the object size (300~mm).}
\label{fig:Phase2_com_result}
\end{figure}

The estimated object parameters from Phase II are summarized in Table~\ref{tab:mass_estimation} and Table~\ref{tab:com_estimation}, where GT, Sim, and Real denote ground truth, simulation, and real-world experiments, respectively. Three independent runs are reported for each configuration. 

In simulation, small measurement noise is introduced to the contact wrench signals to reflect realistic sensing conditions. For both configurations, the simulation results closely match the ground truth across all runs, with negligible errors in both mass and CoM location. The mass error remains below $0.1\%$, and the CoM error also remains below $0.1\%$, confirming the correctness of the estimation formulation under near-ideal conditions. 

In the real-world experiments, both the estimated mass and CoM exhibit consistent deviations from the ground truth across runs. The mass is overestimated by approximately $3$-$9\%$. Despite this bias, the estimates remain consistent across repeated runs, indicating that the error is systematic rather than purely random. Similarly, the CoM estimates show an offset from the ground truth but remain tightly clustered within each configuration, indicating low run-to-run variability. Overall, the CoM error remains within a few percent relative to the object dimensions $(300~\mathrm{mm}\times200~\mathrm{mm})$, demonstrating that the method provides reasonably accurate and repeatable parameter estimates in practical settings. The sources of the observed systematic error are discussed further in Section~\ref{sec:discussion}.

To provide a more intuitive visualization of the observed CoM offset, \figref{fig:Phase2_com_result} shows the CoM estimation result for Configuration~2, Run~1, represented as the inferred additional-mass location. This visualization converts the estimated CoM of the combined box-additional-mass system into the corresponding location of the internal added mass, making the effect of the CoM estimation error easier to interpret. 

Across three simulation trials per configuration, lifting and measurement collection took $2.52477 \pm 0.00012\,\mathrm{s}$ for Configuration~1 and $2.53353 \pm 0.00006\,\mathrm{s}$ for Configuration~2, including the prescribed $2\,\mathrm{s}$ hold. Values are reported as mean $\pm$ sample standard deviation. No variation in completion time was observed at the $1\,\mathrm{ms}$ logging resolution. For the available real-robot recordings, the corresponding durations were approximately $45.254 \pm 0.353\,\mathrm{s}$ for Configuration~1 and $43.940 \pm 0.134\,\mathrm{s}$ for Configuration~2. These real-robot durations were calculated by adding the nominal $5\,\mathrm{s}$ hold to the recorded lifting time. These durations exclude parameter-estimation computation. The estimator solves linear least-squares problems and requires only a few milliseconds with a mean of $1.94\,\mathrm{ms}$ and $1.72\,\mathrm{ms}$ for Configurations~1 and~2, respectively.

The real-robot execution took longer because the commanded lift was slower and the measurement hold was longer. The reference trajectory was assigned a duration of $120\,\mathrm{s}$ and stopped early when both end effectors had risen by more than $h_{\mathrm{lift}}=15\,\mathrm{mm}$, following \eqref{eq:liftoff_condition}. This occurred after approximately $39$--$40\,\mathrm{s}$. The simulation used a different motion schedule and a lift threshold of $10\,\mathrm{mm}$. These timings therefore reflect the execution settings used in each environment, rather than a direct comparison of estimation speed.

The estimated mass and CoM are then used in Phase~III to determine the contact forces and moments required to support and transport the box.

\subsection{Phase III Results (Contact Force and Moment Optimization via Second-Order Cone Programming)}
\label{sec:phase3_result}

\begin{figure*}[!t]
    \centering
        \subfloat[]{
        \includegraphics[width=0.46\textwidth]{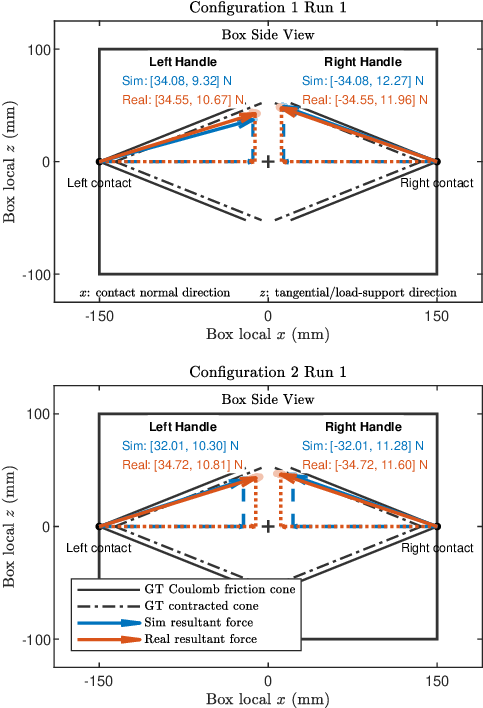}
        \label{fig:result_phase3_friction_cone}
        }
    \hfill
        \subfloat[]{
        \includegraphics[width=0.49\textwidth]{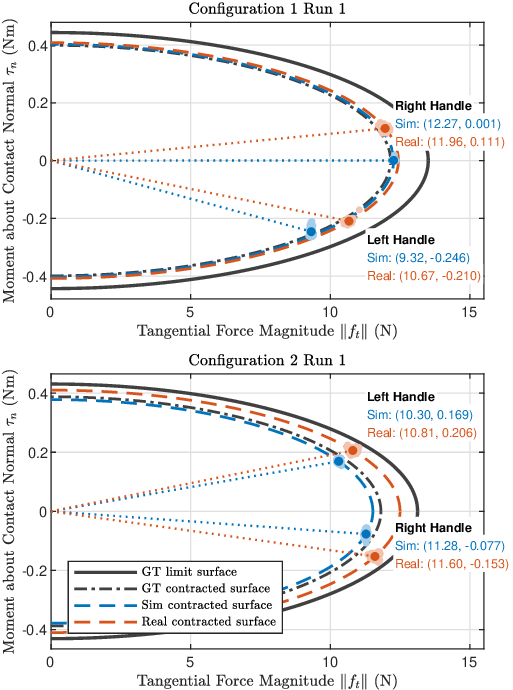}
        \label{fig:result_phase3_limit_surface}
        } 
    \caption{Phase~III SOCP results for Configuration~1 and~2 (both Run~1). (a) Coulomb friction cone defined by \eqref{eq:f_tan_max} and incorporated in the friction limit surface \eqref{eq:limitsurface_ellipsoid}. Contact forces are shown in the local $x$-$z$ plane. Solid and dashed gray lines denote the ground-truth friction cone and its contracted version, respectively. Blue and orange arrows indicate resultant contact forces from simulation and real-world experiments. (b) Friction limit surfaces defined by \eqref{eq:limitsurface_ellipsoid} and optimized tangential force-torsional moment samples. Solid curves denote ground-truth limit surfaces, while dashed curves denote contracted feasible regions with safety margin $r_s=0.10$.}
    \label{fig:phase3_friction_results}
\end{figure*}

The optimized contact wrenches obtained from solving \eqref{eq:phase3_socp_reduced} are visualized in \figref{fig:phase3_friction_results} for both configurations, shown for a representative run (Run~1). The results from the remaining runs exhibit similar trends and are therefore omitted for clarity. \figref{fig:phase3_friction_results} evaluates the optimized wrenches from two complementary perspectives: the Coulomb friction cone in \figref{fig:phase3_friction_results}(a), which shows the contact force feasibility, and the friction limit surface in \figref{fig:phase3_friction_results}(b), which shows the coupling between tangential force and torsional moment. 

As shown in \figref{fig:phase3_friction_results}(a), the contact force vectors are plotted in the local $x$-$z$ plane, where the $x$-direction corresponds to the contact normal direction and the $z$-direction corresponds to the load-support direction. The solid gray lines denote the nominal Coulomb friction cone, while the dashed gray lines denote the contracted cone associated with the safety margin $r_s=0.10$. For both configurations, the optimized contact forces remain within the contracted cones, confirming that the load-supporting tangential forces are feasible for the generated squeezing forces. The contact force distribution is physically consistent with the estimated CoM offset. In both configurations, the tangential force component at the right handle is larger than that at the left handle. This occurs because the CoM is offset along the $+x_{\mathrm{local}}$-direction (see Table~\ref{tab:com_estimation}), requiring an asymmetric distribution of load-supporting tangential forces to satisfy moment equilibrium. At the same time, the normal force components at the two contacts act as opposing squeezing forces with equal magnitude and opposite direction, consistent with force equilibrium in the contact-normal direction. 

The corresponding tangential force-torsional moment pairs are shown in \figref{fig:phase3_friction_results}(b). The horizontal axis represents the tangential force magnitude $\|\mathbf f_t\|$, and the vertical axis represents the torsional moment $\tau_n$ about the contact normal. The solid curves denote the nominal friction limit surfaces, while the dashed curves denote the contracted feasible regions with safety margin $r_s=0.10$. The optimized wrench samples lie inside the contracted feasible regions for both configurations, indicating that the SOCP solution satisfies the friction limit surface constraints with the prescribed safety margin. The smaller torsional moment magnitude at the right handle is also consistent with the friction-limit-surface coupling, as defined in \eqref{eq:limitsurface_ellipsoid}. Since the right handle carries a larger tangential force, it uses a larger portion of the available friction budget. As a result, the admissible torsional moment about the contact normal is reduced at that contact, and the optimized solution assigns a smaller $|\tau_n|$ to the right handle. This illustrates why tangential force and torsional moment cannot be selected independently when the friction limit surface is considered. Across Configuration~1 and Configuration~2, the sign of the torsional moment at each handle reverses. This is consistent with the change in CoM offset along the $y_{\mathrm{local}}$-direction: Configuration~1 has a positive $y_{\mathrm{local}}$ offset, whereas Configuration~2 has a negative $y_{\mathrm{local}}$ offset. The reversal indicates that the required torsional compensation changes direction when the object mass distribution is changed. 

Differences between the simulation and real-world results are mainly due to the mass and CoM estimation errors from Phase~II. Nevertheless, the optimized contact wrenches remain inside the friction-feasible regions and follow the expected force-moment distribution for each CoM configuration. 

The contact-wrench optimization was solved once before transport to determine the desired contact forces and moments. The resulting wrenches were then used as the desired contact wrenches for the execution stage in Sec.~\ref{sec:execution_result}, without repeatedly solving the optimization problem in the control loop.

\subsection{Execution Results}
\label{sec:execution_result}

\begin{figure}[!t]
    \centering
        \subfloat[]{\includegraphics[width=\linewidth]{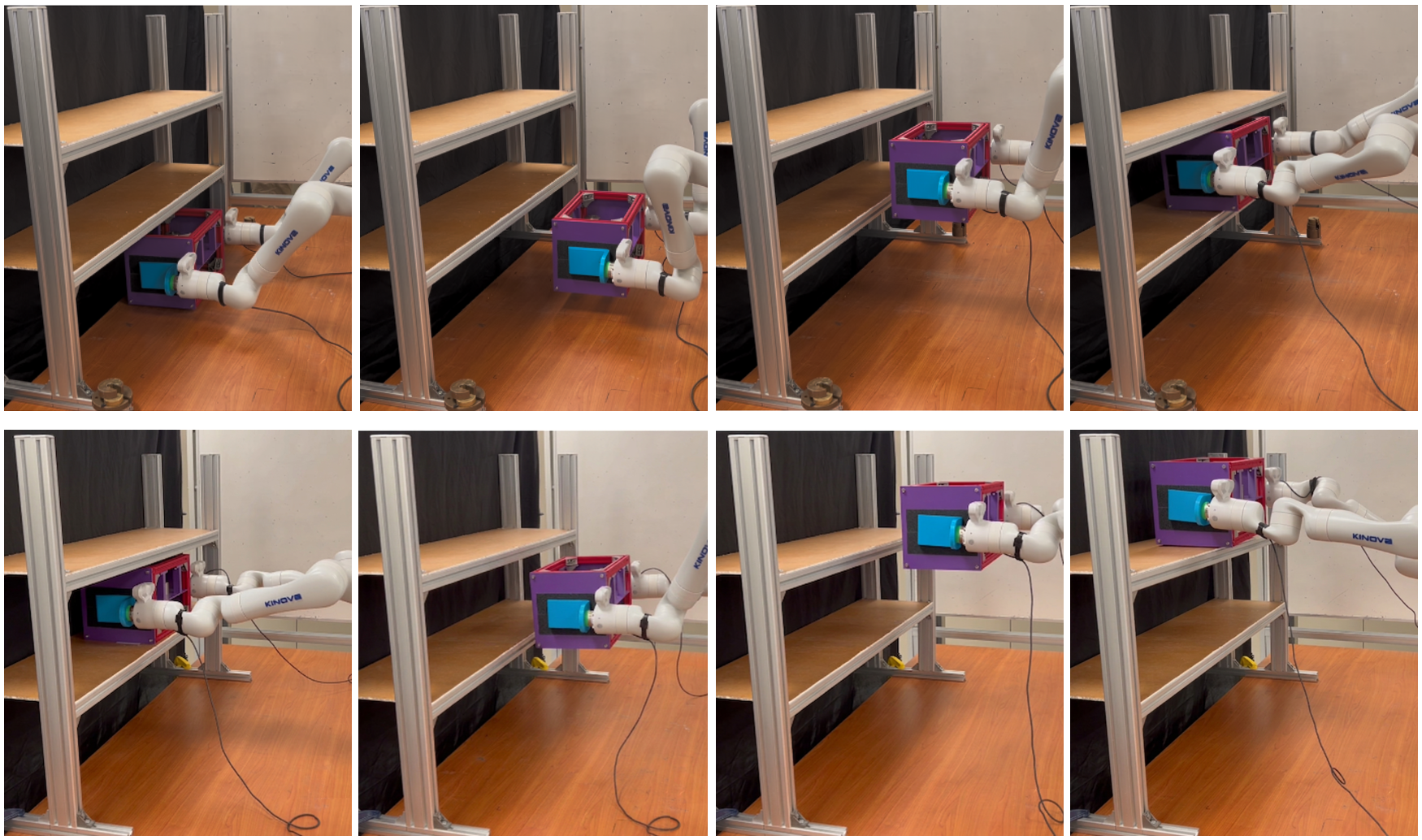}
        \label{fig:Phase4_trajectory_photos}}
    \hfil
        \subfloat[]{\includegraphics[width=0.8\linewidth]{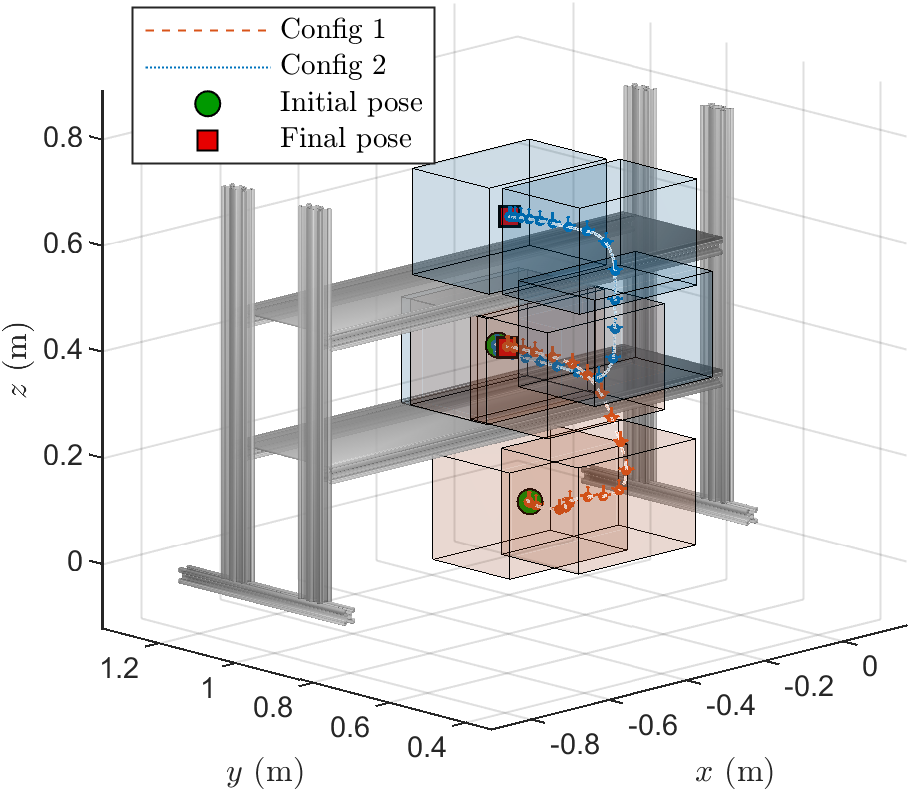}
        \label{fig:result_phase4_traj}}
    \caption{Execution results for the full pipeline. (a) Representative image sequences for Configuration~1 Run~1 (top) and Configuration~2 Run~1 (bottom). (b) Reconstructed 3D box motion in the shelf environment for the two configurations. In both cases, the motion is executed using the optimized contact wrench from Phase~III based on the online CoM estimate from Phase~II.}
    \label{fig:phase4_results}
\end{figure}

The full pipeline is evaluated on the dual-arm system described in Sec.~\ref{sec:experiment_setup}. \figref{fig:phase4_results}(a) shows representative execution sequences for the two CoM configurations, and \figref{fig:phase4_results}(b) shows the corresponding reconstructed 3D box trajectories. As discussed in Sec.~\ref{sec:phase1result}, Configuration~1 uses the refined trajectory from Phase~I, whereas Configuration~2 does not require trajectory refinement. In both cases, execution uses the optimized contact wrench from Phase~III based on the online CoM estimates from Phase~II. Each configuration is repeated three times, and all runs show consistent behavior.

\subsection{Comparison with Existing Methods}
\label{sec:comparison_with_existing_methods}

\begin{table*}[t]
\centering
\small
\caption{Simulation comparison of the proposed method with the BB baseline and two Bi-Touch policies during box transport and holding. Peak angle and mean total squeezing force are evaluated over $3.5$--$7\,\mathrm{s}$, and maximum position error during the final hold is evaluated over $6$--$7\,\mathrm{s}$.}
\label{tab:baseline_comparison}
\setlength{\tabcolsep}{7pt}
\begin{tabular}{llrrrl}
\hline
Method & \shortstack{CoM\\configuration} & \shortstack{Peak angle\\$\theta_{\max}$ (deg)} & \shortstack{Mean total squeezing\\$\overline f_{\mathrm{sq}}$ (N)} & \shortstack{Hold max error\\$e_{\mathrm{hold}}$ (mm)} & Outcome \\
\hline
Proposed & Centered & 0.0001 & 59.9 & 0.117 & Pass \\
BB & Centered & 0.0001 & 75.4 & 0.212 & Pass \\
Bi-Touch broad & Centered & 1.4183 & 66.3 & 11.998 & Pass \\
Bi-Touch narrow & Centered & 0.4515 & 60.3 & 12.264 & Pass \\
\hline
Proposed & Configuration 1 & 0.7493 & 68.1 & 0.203 & Pass \\
BB & Configuration 1 & 0.7493 & 75.4 & 0.226 & Pass \\
Bi-Touch broad & Configuration 1 & 3.0668 & 68.0 & 12.606 & Pass \\
Bi-Touch narrow & Configuration 1 & 23.8388 & 58.5 & 48.754 & Fail \\
\hline
Proposed & Configuration 2 & 0.6411 & 64.0 & 0.301 & Pass \\
BB & Configuration 2 & 0.6411 & 75.4 & 0.213 & Pass \\
Bi-Touch broad & Configuration 2 & 1.0998 & 67.6 & 12.385 & Pass \\
Bi-Touch narrow & Configuration 2 & 23.8749 & 59.8 & 49.896 & Fail \\
\hline
\end{tabular}
\end{table*}

\begin{figure*}
    \centering
    \includegraphics[width=\textwidth]{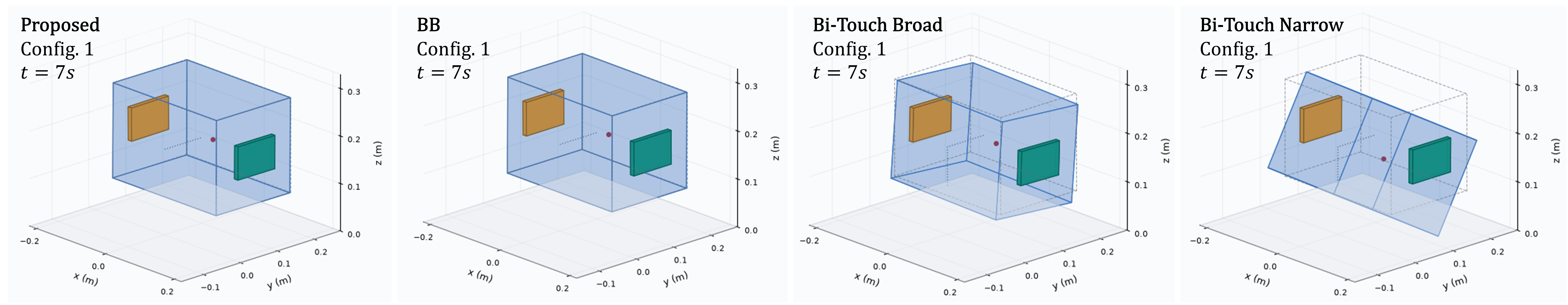}
    \caption{Final box poses in simulation for Configuration~1 at $t=7\,\mathrm{s}$ using the proposed method, BB, Bi-Touch broad and Bi-Touch narrow, from left to right. The proposed method and BB maintain the box close to the desired upright pose. Bi-Touch broad shows a small pose deviation, while Bi-Touch narrow shows heavy tilting.}
     \label{fig:comparison_config1_end}
\end{figure*}

The proposed method is compared with one existing model-based approach and one existing learning-based approach for dual-arm box manipulation. The model-based comparison uses the contact-wrench allocator of Bombile and Billard (BB)~\cite{bombile2022dual}. The learning-based comparison uses two policies adapted from Bi-Touch~\cite{lin2023bi}. For BB, the proposed Phase~II parameter-estimation procedure is retained, and the BB allocator replaces the Phase~III Contact Wrench Optimization via SOCP. Bi-Touch instead learns to control the handles from tactile feedback. The comparison examines the squeezing force and object motion during transport and holding after the initial lift.

To provide a comparison of these methods, a simple task is introduced. The task starts with squeezing the box over $0$--$0.8\,\mathrm{s}$, followed by an $80\,\mathrm{mm}$ lift over $0.8$--$2.8\,\mathrm{s}$ and a stationary hold until $3.5\,\mathrm{s}$. The box is then moved horizontally by $100\,\mathrm{mm}$ over $3.5$--$6\,\mathrm{s}$ and held at the target pose until $7\,\mathrm{s}$.

All methods are evaluated in the same simulation, using the same box, same horizontal contact patches of $100\times70\,\mathrm{mm}$, same friction coefficient $\mu=0.4$, same motion reference and actuator limits. The translational and rotational impedance stiffnesses are $1000\,\mathrm{N/m}$ and $10\,\mathrm{Nm/rad}$, respectively. The simulation represents the box and two moving handles. The shelf-avoidance task of Phase~I and individual robot-joint dynamics are not included.

The box mass is fixed at $2.2\,\mathrm{kg}$. Three CoM configurations are considered: Centered, Configuration~1 and Configuration~2, with in-plane coordinates $(r_x,r_y)$ of $(0,0)$, $(20.50,11.40)$ and $(6.80,-11.40)\,\mathrm{mm}$, respectively. The two eccentric loading configurations follow Tables~\ref{tab:mass_estimation} and~\ref{tab:com_estimation}, with Centered added as a reference case. For the proposed method and BB, mass and CoM are estimated from simulated contact-force and moment measurements after a common preliminary lift, as described in Phase~II (Sec.~\ref{sec:methods_phase2}). Both methods then receive the same estimates and begin transport from the same box and handle state. This keeps the estimation stage fixed when comparing the two wrench allocators.

BB uses quadratic programming to select the contact forces and moments. Its objective balances two terms: the difference between the resulting and desired object wrench, and the difference between the contact wrenches and a reference wrench. The authors' solver is retained, with the friction and contact-geometry settings adapted to the present box. BB uses a linear approximation of the friction cone and separate bounds on torsional and bending moments, while this paper's proposed method uses the coupled friction limit surface and safety margin $r_s=0.10$ described in Sec.~\ref{sec:methods_phase3}. Both methods receive the same desired object wrench and execute their contact-wrench commands through the same impedance controller and normal-force feedback. The BB comparison therefore evaluates its adapted wrench allocator, rather than its original fast-grabbing and tossing task \cite{bombile2022dual}.

The Bi-Touch comparison uses two policies: a broad policy and a narrow policy, trained with different ranges of CoM positions. The broad policy samples $r_x$ uniformly from $[-25,25]\,\mathrm{mm}$ and $r_y$ from $[-15,15]\,\mathrm{mm}$, so both eccentric Configuration~1 and Configuration~2 lie within its training range. The narrow policy samples both coordinates from $[-5,5]\,\mathrm{mm}$, placing the eccentric configurations outside its training range. Centered lies within both ranges. The mass remains fixed and $r_z=0$. Both policies use Proximal Policy Optimization (PPO), retaining the authors' learning algorithm and tactile-image feature extractor. Each policy receives two simulated tactile images together with handle poses and velocities, the motion reference, task time and current handle offsets. It produces bounded handle-pose adjustments at $10\,\mathrm{Hz}$, without explicit mass or CoM inputs. The observation and action interfaces are adapted to the box task, and the reward encourages tracking and upright motion while penalizing squeezing and force-limit violations. Both policies were trained with the same budget of approximately $100,000$ environment steps. Model checkpoints were selected using separate validation cases, and policy weights remained fixed during testing.

The reported motion and force metrics cover $t_0=3.5\,\mathrm{s}$ to $t_h=6\,\mathrm{s}$, when the box is transported horizontally, and from $t_h=6\,\mathrm{s}$ to $t_f=7\,\mathrm{s}$, when the box is held at the target. Using the same time window avoids comparing force averages over different parts of the task. Let $\hat f_{n,i}(t)=\mathbf n_i^\top\hat{\mathbf f}_i(t)$ be the measured signed normal force, following \eqref{eq:f_ni_andf_ti}, so that $-\hat f_{n,i}(t)$ is the compressive magnitude. Let $\theta(t)$ denote the box's angular deviation from the desired upright orientation, and let $\mathbf p_{\mathrm{box}}(t)$ and $\mathbf p_{\mathrm{box}}^{\mathrm{ref}}(t)$ denote its measured and desired positions. The evaluation metrics are
\begin{align}
    \theta_{\max}&=\max_{t\in[t_0,t_f]}\theta(t),\\
    \overline f_{\mathrm{sq}}&=\frac{1}{t_f-t_0}\int_{t_0}^{t_f}\sum_{i\in\{L,R\}}[-\hat f_{n,i}(t)]\,dt,\\
    e_{\mathrm{hold}}&=\max_{t\in[t_h,t_f]}\left\|\mathbf p_{\mathrm{box}}(t)-\mathbf p_{\mathrm{box}}^{\mathrm{ref}}(t)\right\|_2.
\end{align}
Here, $\theta_{\max}$ is the peak angle in degrees, $\overline f_{\mathrm{sq}}$ is the mean total squeezing force in newtons, and $e_{\mathrm{hold}}$ is the maximum position error during holding, reported in millimetres. A passing segment requires hold error below or equal to $15\,\mathrm{mm}$, peak angle below or equal to $5^\circ$, normal force no greater than $65\,\mathrm{N}$ per handle, relative tangential contact displacement no greater than $40\,\mathrm{mm}$, and no drop. These metrics evaluate the response after estimation. They do not measure Phase~II estimation accuracy or the preceding lift. Bi-Touch reaches this interval through its own execution, so its starting state can differ from that of the two model-based methods.

Table~\ref{tab:baseline_comparison} shows that the proposed method and BB complete all three configurations. However, less squeezing force is required by the proposed method than by BB. Both methods maintain small peak angles and sub-millimetre hold errors, indicating that the proposed method reduces contact normal force while maintaining stable transport and accurate holding. Their nearly identical peak angles in the eccentric cases include the same initial post-lift deviation.

Bi-Touch broad also completes all three configurations, with mean squeezing forces of a similar magnitude to those of the proposed method. However, its larger peak angles and hold errors indicate greater box rotation and less accurate positioning during holding. The proposed method therefore maintains more accurate object motion with comparable squeezing force in these cases. Bi-Touch narrow succeeds in the centered case but fails in both eccentric configurations, which lie outside its CoM training range. These failures involve excessive tilt, position error and force-limit violations. The lower mean squeezing force in these cases does not represent successful force reduction. The Bi-Touch results are based on one training seed per policy, with differences between the contact models used for training and evaluation. The observed performance therefore reflects these particular policies and test conditions.

\figref{fig:comparison_config1_end} shows the corresponding final box poses at the end of the holding phase for Configuration~1. The proposed method and BB keep the box close to the desired pose, while Bi-Touch broad shows a small pose deviation and Bi-Touch narrow exhibits pronounced tilting.

\subsection{Ablation Study}
\label{sec:ablation_study}

\begin{table}
\centering
\footnotesize \caption{Ablation study evaluating the contribution of each phase in the proposed pipeline.}
\label{tab:ablation_study}
\setlength{\tabcolsep}{4pt}
\renewcommand{\arraystretch}{1}
\begin{tabular}{ | m{1.5cm} | m{1.0cm} | m{1.4cm} | m{1.5cm} | m{1.9cm} | }
\hline
\textbf{Method Variant} & \textbf{Collision} & \textbf{Orientation Deviation} & \textbf{Wrench Quality} & \textbf{Notes} \\
\hline
Full pipeline & No & No & Optimal & Uses Phase I-III \\
\hline
w/o Phase I & Yes & -  & - & Trajectory may collide with environment \\
\hline
w/o Phase II & No & Yes (tilted) & Suboptimal & Incorrect CoM estimate leads to wrong wrench \\
\hline
w/o Phase III & No & Yes (tilted) & Inefficient, unbalanced, excessive squeezing & No optimal wrench \\
\hline
\end{tabular}
\end{table}

An ablation study is conducted to evaluate the contribution of each phase in the proposed pipeline, as summarized in Table~\ref{tab:ablation_study}. The full pipeline achieves collision-free manipulation while maintaining the desired object orientation by combining trajectory refinement (Phase~I), online CoM estimation (Phase~II), and contact wrench optimization (Phase~III).

When Phase I is removed, the system relies on the nominal trajectory, which may violate environmental constraints and result in collisions in confined settings, as also can be observed in \figref{fig:result_phase1_combined}.

When Phase II is removed, the system relies on a nominal assumption that the center of mass is located at the geometric center of the object, causing the computed contact wrench to be inconsistent with the true mass distribution. As a result, the applied forces and moments do not properly balance the object, leading to noticeable orientation deviation during manipulation.

When Phase III is removed, no optimal contact wrench is available, and the system instead relies on a naive symmetric force assignment, where both arms apply equal normal forces, share the tangential load equally, and no contact moment is commanded. As a result, the forces and moments are not properly adapted to the object's mass distribution. This leads to inefficient and unbalanced force application, often requiring excessive squeezing to maintain the grasp, and may still result in tilted motion due to the lack of appropriate moment compensation.

These results highlight that each phase plays a distinct and complementary role: Phase I ensures collision-free motion, Phase II provides accurate object parameter estimation, and Phase III enables physically consistent and efficient force execution.

\section{Discussion}
\label{sec:discussion}

The proposed framework connects trajectory refinement, post-lift estimation of unknown object loading, and friction-constrained wrench execution for quasi-static dual-arm box manipulation. Its contribution lies in this integration and its application to selecting contact forces and torsional moments from the estimated mass and CoM. The underlying least-squares estimation, convex optimization, and friction models are established tools.

The simulation comparisons help assess the practical benefit of this formulation. With the same parameter estimates and execution controller, the proposed method requires less squeezing than the adapted BB allocator while both maintain small orientation deviations and holding errors. Compared with the task-adapted Bi-Touch broad policy, it achieves smaller motion errors with comparable squeezing forces. These findings apply to the tested adaptations and conditions; the Bi-Touch comparison is additionally limited by one training seed per policy and differences between the training and evaluation contact models.

\paragraph{Slip avoidance and reduced squeezing as two facets of optimization formulation}
The optimization addresses friction feasibility and contact effort within a single formulation. Phase II supplies the estimated loading, while Phase III selects contact forces and torsional moments that satisfy equilibrium and friction constraints with minimum weighted effort. This provides a systematic way to limit unnecessary squeezing while maintaining modeled contact feasibility. The physical experiments demonstrate slip-free and drop-free manipulation under the tested conditions, although execution still depends on the accuracy of the contact model and wrench tracking.

\paragraph{Role of the safety margin}
The safety margin $r_s$ is defined through a Pontryagin difference in normalized friction coordinates. The contracted surface leaves room for bounded tangential-force and torsional-moment perturbations at a fixed normal force while preserving the SOCP structure. Increasing $r_s$ reduces the available friction capacity for a given normal force and may therefore require greater squeezing to support the same load. This margin provides a tunable degree of conservatism within the assumed contact model.

\paragraph{Influence of the cost function on the force-moment} distribution
The force-moment distribution in Phase~III depends on the contact model. The tangential components of the contact moment are constrained to zero by \eqref{eq:no_bending_moment}. Differences in supporting forces balance the gravitational moment perpendicular to the line joining the contacts, while contact torsional moments balance its component along that line. Equilibrium constrains the combined effect of these wrenches, but their distribution between the contacts also depends on friction feasibility and the weighting matrix. The characteristic length $l_c$ in \eqref{eq:contact_weight_matrix} sets the relative penalty on contact moments and can therefore affect the allocation even when bending moments are excluded.

If the constraint in \eqref{eq:no_bending_moment} were removed, tangential contact moments would also become admissible. The optimizer could then distribute the required moment balance between force asymmetry and contact moments, with the tradeoff influenced by the moment weighting through $l_c$. Both cases satisfy the same equilibrium constraints, but correspond to different contact models and force-moment distributions.

\paragraph{Simulation-to-Real Setup Consistency in Phase I}
Phase I trajectory refinement is performed in simulation, and its effectiveness depends on how closely the simulated environment matches the real-world setup. In practice, even small differences in object geometry, contact properties, or shelf configuration can lead to residual contact during execution. In addition, the current formulation assumes that the object geometry (e.g., box dimensions) is known beforehand. In more general settings, this information would need to be obtained through perception before trajectory refinement can be applied.

\paragraph{Systematic Error in CoM Estimation in Phase II}
Experimental results show a consistent bias in the estimated center of mass. Possible sources include sensor noise, robot kinematic inaccuracies, and modeling mismatch arising from the handle contact locations not being perfectly aligned with the geometric center of the box as assumed in the model \cite{turlapati2024identification}. In practice, such deviations introduce moment offsets that directly affect the estimation, leading to systematic rather than random error.

Overall, the results indicate that the framework can successfully execute stable lifting with online estimation of unknown object properties, while highlighting key areas for improvement.

\section{Conclusion}

This work presented a friction-aware dual-arm box-manipulation framework for objects with unknown mass and center of mass. The approach combines offline trajectory refinement, online parameter estimation, and optimization-based contact wrench computation under friction constraints. The optimized wrenches are executed through impedance control on a real dual-arm robotic system.

Experiments show that the proposed method lifts objects under different CoM configurations while avoiding slip, object drop, and excessive squeezing. The results demonstrate the importance of online CoM estimation and explicit force-moment optimization for efficient frictional manipulation.

In the simulation comparison with existing methods, the proposed method required less squeezing than BB while maintaining similarly small orientation deviations and holding errors. Compared with the task-adapted Bi-Touch broad policy, it achieved smaller orientation deviations and holding errors with comparable squeezing forces. These results demonstrate that the proposed method maintains stable and accurate manipulation while limiting unnecessary squeezing under the tested conditions.

Future work will focus on improving robustness to modeling mismatch, integrating perception for object geometry estimation, and extending the framework to dynamic manipulation tasks and richer contact interactions.

\appendix

\section{Safety Margin for Friction Constraints}
\label{app:safety_margin_rs}

\begin{figure}[!t]
    \centering
        \subfloat[]{\includegraphics[width=0.6\linewidth]{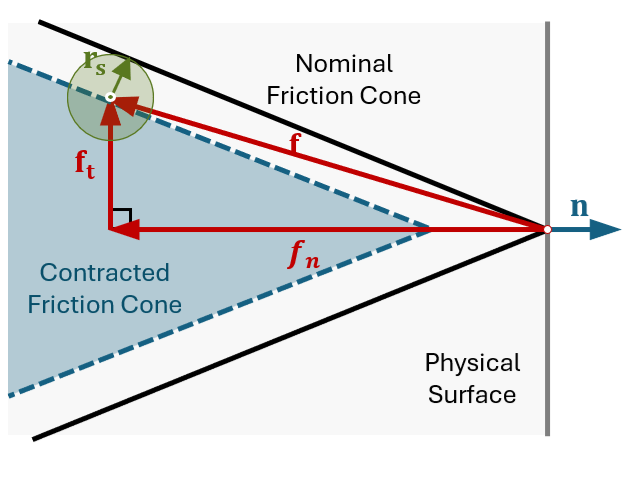}
        \label{fig:Appendix_Friciton_Cone}}
    \hfil
        \subfloat[]{\includegraphics[width=0.48\linewidth]{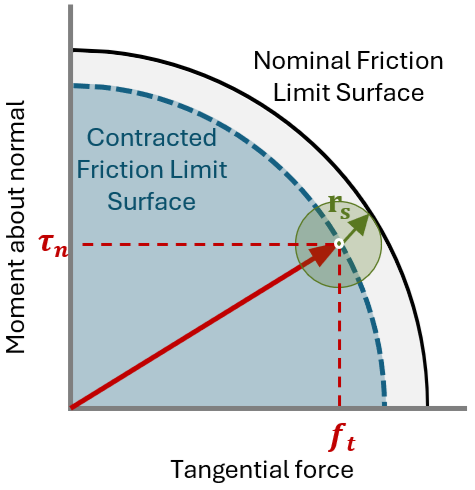}
        \label{fig:Appendix_Friciton_Limit_Surface}}
    \caption{Illustration of the safety margin $r_s$ for (a) the friction cone and (b) the friction limit surface. The green ball represents the admissible perturbation set used to contract the nominal friction constraints.}
    \label{fig:appendix_safetymargin}
\end{figure}

A normalized safety margin $r_s\in[0,1)$ is introduced by contracting the admissible friction region. For a fixed compressive normal force $f_n<0$, the normalized tangential force and torsional moment are defined as
\begin{equation}
    \bar{\mathbf f}_t=\frac{\mathbf f_t}{\mu(-f_n)},
    \qquad
    \bar{\tau}_n=\frac{\tau_n}{\tau_n^{\max}},
\end{equation}
where $\tau_n^{\max}=\mu(-f_n)R_{\mathrm{eff}}$. 

Let $\bar{\mathbf w}$ denote either $\bar{\mathbf f}_t$ for the friction-cone cross-section or $[\bar{\mathbf f}_t^\top\;\bar{\tau}_n]^\top$ for the coupled friction region. In both cases, the nominal normalized feasible set is
\begin{equation}
    \mathcal F=\left\{\bar{\mathbf w}\mid\|\bar{\mathbf w}\|_2\le1\right\}.
\end{equation}
The contracted feasible set is defined as
\begin{equation}
    \mathcal F_{r_s}=\mathcal F\ominus\mathbb B_{r_s},
\end{equation}
where $\ominus$ denotes the Pontryagin difference \cite{cotorruelo2022pontryagin,schneider2013convex}, and
\begin{equation}
    \mathbb B_{r_s}
    =\left\{\Delta\bar{\mathbf w}\mid
    \|\Delta\bar{\mathbf w}\|_2\le r_s\right\}
\end{equation}
is a closed ball of admissible perturbations in normalized coordinates. Equivalently,
\begin{equation}
    \mathcal F_{r_s}
    =\left\{\bar{\mathbf w}\mid
    \bar{\mathbf w}+\mathbb B_{r_s}\subseteq\mathcal F\right\}
    =\left\{\bar{\mathbf w}\mid
    \|\bar{\mathbf w}\|_2\le1-r_s\right\}.
\end{equation}

Thus, every point in the contracted region remains within the nominal region under any normalized perturbation of magnitude at most $r_s$. The green ball in \figref{fig:appendix_safetymargin} illustrates this interpretation. The perturbations concern tangential force and torsional moment at a fixed normal force. They do not include changes in the normal force itself.

Returning to physical coordinates gives the factor $(1-r_s)$ used in the Phase~III friction constraints. For the coupled model, this yields the SOC constraint in \eqref{eq:friction_soc_explicit}. Although the normalization above requires $f_n<0$, the SOC form also remains valid at $f_n=0$, where it requires both the tangential force and torsional moment to be zero. When $r_s=0$, the nominal constraints are recovered. Increasing $r_s$ reduces the admissible tangential force and torsional moment for a given normal force, providing a larger margin within the assumed friction model.

\section{Derivation of Torsional Friction Limit}
\label{app:torsional_friction}

\subsection{General derivation}
Following \cite{howe1996practical}, consider a rigid contact with a fixed contact area $A$ under pure twist about the contact normal. Let $p_{\mathrm c}(\boldsymbol\rho)$ denote the normal pressure distribution over the patch, where $\boldsymbol\rho\in\mathbb{R}^2$ is the in-plane position vector measured from the center of rotation. Under Coulomb friction, the maximum admissible torsional moment about the normal direction can be written as
\begin{equation} \label{eq:tau_max_howe}
    \tau_n^{\max} = \mu \int_{A} p_{\mathrm c}(\boldsymbol\rho) \, \|\boldsymbol\rho\| \, dA,
\end{equation}
where $\mu$ is the friction coefficient. Using the signed normal force convention from Phase~III, the compressive magnitude is
\begin{equation}
    -f_n \triangleq \int_{A} p_{\mathrm c}(\boldsymbol\rho)\, dA,
\end{equation}
For $f_n<0$, the \emph{effective contact radius} is defined as
\begin{equation} \label{eq:reff_def}
    R_{\mathrm{eff}} \triangleq \frac{1}{-f_n}\int_{A} p_{\mathrm c}(\boldsymbol\rho)\,\|\boldsymbol\rho\|\, dA,
\end{equation}
so that \eqref{eq:tau_max_howe} becomes the compact expression
\begin{equation} \label{eq:tau_max_reff}
    \tau_n^{\max} = \mu(-f_n)R_{\mathrm{eff}}.
\end{equation}
The value of $R_{\mathrm{eff}}$ depends on the contact geometry and the assumed pressure distribution $p_{\mathrm c}(\boldsymbol\rho)$.

\subsection{Evaluation for uniform pressure over a rectangular patch}
Assume the contact patch is a rectangle of size $a \times b$ (in meters), centered at the origin, with a uniform pressure distribution
\begin{equation}
    p_{\mathrm c}(\boldsymbol\rho) = \frac{-f_n}{ab}, \qquad \boldsymbol\rho=(x,y)^\top,\quad (x,y)\in\left[-\frac{a}{2},\frac{a}{2}\right]\times\left[-\frac{b}{2},\frac{b}{2}\right].
\end{equation}
Substituting into \eqref{eq:reff_def} yields
\begin{equation} \label{eq:reff_rect_integral}
    R_{\mathrm{eff}}
    = \frac{1}{ab}
    \int_{-a/2}^{a/2}\int_{-b/2}^{b/2}\sqrt{x^2+y^2}\,dy\,dx.
\end{equation}
Evaluating \eqref{eq:reff_rect_integral} gives the closed-form expression
\begin{equation} \label{eq:reff_rect_closedform}
    R_{\mathrm{eff}}
    =
    \frac{\sqrt{a^2+b^2}}{6}
    +
    \frac{a^2}{12b}\,\mathrm{asinh}\!\left(\frac{b}{a}\right)
    +
    \frac{b^2}{12a}\,\mathrm{asinh}\!\left(\frac{a}{b}\right),
\end{equation}
where $\mathrm{asinh}(\cdot)$ denotes the inverse hyperbolic sine. For the contact geometry used in this work, with $a=100$~mm and $b=70$~mm, \eqref{eq:reff_rect_closedform} yields $R_{\mathrm{eff}} \approx 0.0328~\text{m}.$

\section*{Declaration of generative AI and AI-assisted technologies in the manuscript preparation process}
During the preparation of this work the authors used OpenAI's ChatGPT in order to  assist with language refinement and readability of selected manuscript sections. After using this tool/service, the authors reviewed and edited the content as needed and take full responsibility for the content of the published article.

\bibliographystyle{elsarticle-num}
\bibliography{main}
\end{document}